\documentclass{article}
\usepackage[dvipdfmx]{graphicx}
\usepackage{arxiv}
\usepackage{authblk}

\usepackage{amsmath,amssymb,amsfonts,amsthm,mathtools}
\usepackage{mathrsfs}%
\usepackage[utf8]{inputenc}
\usepackage{hyperref}
\usepackage{enumitem}

\usepackage{xcolor}%
\usepackage{textcomp}%

\usepackage{multirow}
\usepackage{subfig}
\usepackage{array}
\usepackage{booktabs}
\usepackage{threeparttable}
\usepackage{colortbl}

\theoremstyle{thmstyleone}%
\theoremstyle{thmstyletwo}%

\theoremstyle{thmstylethree}%

\newcolumntype{P}[1]{>{\centering\arraybackslash}m{#1}}

\begin{document}

\newcommand{\myPaperShortTitle}{Cross-Block Conditioning in DBMs}
\newcommand{\myPaperTitle}{
Cross-Block Conditioning in Deep Boltzmann Machines for Statistical Data Fusion
}
\title{\myPaperTitle}
\date{}


\renewcommand\Authfont{\bfseries}
\setlength{\affilsep}{0em}
\newbox{\orcid}\sbox{\orcid}{\includegraphics[scale=0.06]{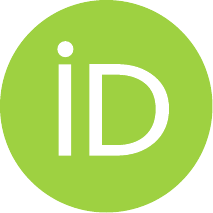}} 
\author[1]{%
	\href{https://orcid.org/0000-0002-4618-6272}{\usebox{\orcid}\hspace{1mm}
	Junichiro Niimi\thanks{\texttt{jniimi@meijo-u.ac.jp}}
	}}
\affil[1]{Meijo University}

\renewcommand{\shorttitle}{\myPaperShortTitle}

\maketitle

\begin{abstract}
Statistical data fusion combines two panels that share a block of covariates but observe disjoint outcome blocks, and in its traditional form no row observes both outcomes at once. That rules out the discriminative criterion one would rather train a Deep Boltzmann Machine with, since multi-prediction training needs ground truth for whatever it holds out. We propose observed-block multi-prediction, which restricts the multi-prediction objective to targets drawn from what each row actually observes. It is well defined for any missingness pattern and reduces to the original criterion when rows are complete. Having a discriminative criterion that survives the setting lets us ask whether the joint model is needed at all, by separating what it contributes into a representation part and an inference part. On two datasets of different kinds, a consumer purchase panel and public-domain census microdata, over grids in sample size and covariate width spanning $40$ cells and $200$ runs per method, almost none of the fine-tuned DBM's advantage comes from generative pre-training, which is confined to the smallest sample size on one dataset and absent on the other. It comes from conditioning on one outcome block when predicting the other. This term amounts to $+0.19$ and $+0.36$ percentage points, is positive in all $40$ cells, never decays as the panels grow (it is flat on one dataset and grows on the other), and requires neither a second hidden layer nor more inference. Against baselines tuned on validation and given the same conditioning, the fine-tuned DBM is the best method in $37$ of the $40$ cells. The imputers that can also condition on the other outcome block mostly lose accuracy when they do, whereas the DBM gains in every cell; since fusion data cannot validate that choice, this is the property that matters. 
\end{abstract}

\renewcommand\thefootnote{\arabic{footnote}}
\setcounter{footnote}{0}

\newcommand{\bccol}[2]{ \multicolumn{#1}{c}{\bfseries{#2}}}

\section{Introduction}
\subsection{Background}

The recent resurgence of energy-based models (EBMs) is unmistakable. Hinton's 2025 Nobel Lecture~\cite{bm} recentered Boltzmann machines~\cite{bm_training} as a foundational paradigm in machine learning. The Energy-Based Transformer~\cite{ebt} recasts next-token prediction as iterative energy descent. LeCun and collaborators continue to advocate the Joint-Embedding Predictive Architecture (JEPA) family~\cite{i-jepa,v-jepa,lecun2022} as an EBM-flavored alternative to autoregressive generative modeling. Survey and tutorial work~\cite{ebm_hitchhiker,ebm_tutorial,ebm_train} together with new theoretical instruments~\cite{ebm_neurips2025} make the revival concrete. Against this backdrop the Deep Boltzmann Machine (DBM)~\cite{dbm}, the classical deep EBM with two or more stacked hidden layers, remains underexplored as an applied tool. Its native compatibility with partial observation is a structural advantage that contemporary EBMs lack, and the setting that exercises that advantage most directly is statistical data fusion in marketing~\cite{kamakura_wedel_1997}: two panels share a common covariate block $X$ but observe disjoint outcome blocks, $Y_A$ in one and $Y_B$ in the
other, and the task is to fill in what each panel did not measure.

\begin{figure}[htb]
   \centering
   \includegraphics[width=0.99\linewidth]{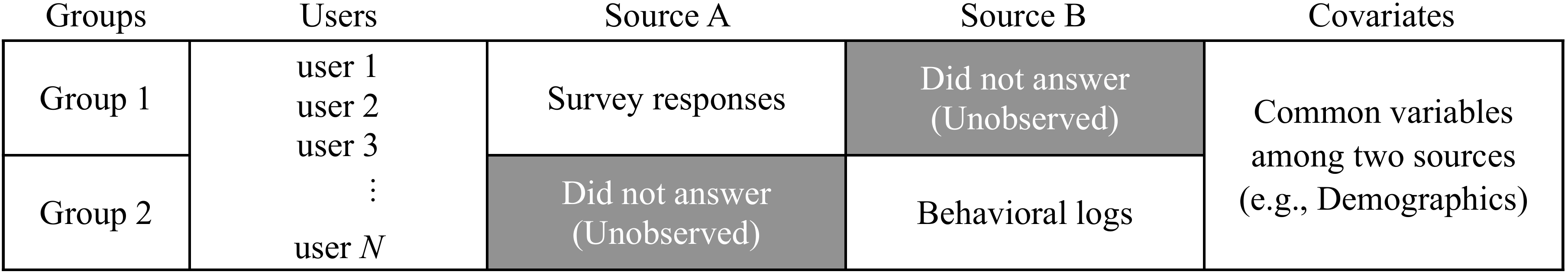}
   \caption{Typical data structure (missing at random, MAR) for statistical data fusion.}\label{fig:datafusion}
\end{figure}

What makes data fusion hard is not the amount of data but its shape. The two outcome blocks are \emph{never observed together} (Figure~\ref{fig:datafusion}). This is a property of the design, not of the sample size: collecting more rows adds more source-$A$ rows and more source-$B$ rows, and never a single row in which $Y_A$ and $Y_B$ appear side by side. Any model of the joint distribution $p(y_A, y_B \mid x)$ therefore has to obtain the association between the two blocks from something other than paired examples, and any model that simply maps $X$ to outcomes has given up on that association by construction.

For a DBM the natural response is to marginalize: missing visible dimensions are summed out of the training objective, which stays well-defined under any observation pattern~\cite{niimi_hoshino_2017}. This generative criterion is available in the fusion regime, but it is not, on its own, a strong predictor. The discriminative alternative --- the Multi-Prediction DBM~\cite{mp-dbm}, which trains the model to be a good inference machine by holding out a random subset of visible dimensions and predicting it --- is the criterion one would rather use, and it is exactly the one that data fusion forbids: its targets must be observed, so when no row observes both blocks, the objective has nothing to score.

This paper closes that gap. We restrict the multi-prediction objective to targets drawn from what each row actually observes. The resulting criterion, which we call \emph{observed-block multi-prediction} (OBMP), is well-defined for any observation pattern, reduces to MP-DBM when every row happens to be complete, and draws gradient from every row in the fusion regime.

Having a discriminative criterion that survives the fusion setting lets us ask the question the two criteria were obscuring, which is the question this paper is really about: \emph{is a joint model necessary here at all?} A discriminative model fitted on the common block is simpler, cheaper, and --- as we will show --- close behind. If what the DBM adds is a better representation, then the difference is one of degree, and a sufficiently flexible discriminative model with enough data should erase it. If instead part of what it adds is unavailable to any model of the form $x \mapsto y$, then the gap has a floor that data cannot lower. Telling those two apart requires separating the contributions rather than reporting their sum, and that is what the experiments below do.

\subsection{Contributions}

\paragraph{A decomposition that separates representation from inference.}
We separate the advantage of the fine-tuned DBM into two parts that can be measured independently: what generative pre-training contributes over a randomly initialized network of the same capacity trained on the same objective and budget, and what conditioning on one outcome block contributes when predicting the other, measured on a single checkpoint under two conditioning sets. On both datasets the second term is the larger by a wide margin: the first is confined to the smallest sample size on Instacart and absent on ACS-housing, while the second is positive in every cell of both grids. The case for the joint model rests on its inference structure, not on generative pre-training as an initializer.

\paragraph{An observed-only multi-prediction criterion.}
We formalize OBMP: a conditioning mask and a target mask constrained to be disjoint and contained in the observed set, scored by a row-normalized cross-entropy on a differentiable mean-field unroll. The criterion is a strict generalization of multi-prediction training to arbitrary missingness, and it is defined in the traditional fusion regime where no training row is complete.

\paragraph{The mechanism does not weaken as data accumulate.}
On two-dimensional grids over sample size $n_{\mathrm{train}}$ and common-block width $k$ ($20$ cells per dataset over five seeds, $100$ runs each), the cross-block term never decays. On Instacart it is flat: between $+0.17$ and $+0.20$\,pp at every sample size from $500$ to $10{,}000$, with no significant trend on either axis. On ACS-housing it grows, from $+0.28$\,pp at $n_{\mathrm{train}}{=}500$ to $+0.45$\,pp at $n_{\mathrm{train}}{=}10{,}000$. The generative pre-training term does not behave this way: on Instacart it is worth $+0.19$\,pp at $n_{\mathrm{train}}{=}500$ and is indistinguishable from zero above $n_{\mathrm{train}}{=}1000$, and on ACS-housing it is positive only at $n_{\mathrm{train}}{=}500$. What the joint model contributes at inference time is therefore not a small-sample effect that more data would erase.

\paragraph{Tuned baselines, the same conditioning, and selection that fusion data allows.}
We compare against baselines spanning marginal predictors, nearest neighbors, chained equations, classical data fusion, factor analysis and deep generative imputation, each tuned on validation and fitted on the same training rows as the model. The joint imputers among them are also given the other outcome block as evidence, as OBMP is, and every setting, OBMP's included, is selected on $X$-only validation, which is all that fusion data allows. OBMP is the best method in $37$ of the $40$ cells. The imputers that can condition on the other block mostly lose accuracy when they do so blindly, whereas OBMP gains in all $40$ cells (\S\ref{sec:res-conditioning}). We read the comparison as evidence that the effect is consistent rather than that it is large: the margins are fractions of a percentage point, on tasks where the distance between a constant predictor and the strongest baseline is itself only $2.4$ and $1.9$ points (\S\ref{sec:disc-size}).

\paragraph{A practical reading.}
The decomposition says which part of the advantage a practitioner should expect to keep. The part that comes from representation shrinks as panels grow, and is the part a tuned discriminative model will eventually match; the part that comes from conditioning on the other outcome block does not shrink, because it is not a matter of what the model learned. It is also a part that a practitioner has to use without being able to check it, since fusion data contain no row on which to validate it, and for OBMP it did not reduce accuracy in any cell we measured.

\section{Related Study}

\subsection{Energy-based models and their resurgence}

An energy-based model defines a probability distribution $p_\theta(x) \propto \exp(-E_\theta(x))$ over a configuration space through a learned scalar energy $E_\theta$. The framework is by construction more flexible than autoregressive likelihood models --- any non-negative function of $x$ is a valid unnormalized density --- at the price of an intractable partition function whose handling is the central technical concern of the field~\cite{ebm_tutorial,ebm_train,ebm_hitchhiker}.

Three concurrent developments have brought EBMs back to the centre of the deep-learning research agenda. First, Hinton's 2025 Nobel Lecture~\cite{bm} explicitly recentered the Boltzmann machine~\cite{bm_training} as a foundational paradigm, foregrounding the maximum-likelihood gradient with its positive and negative phases as a unifying principle. Second, the Energy-Based Transformer~\cite{ebt} reframes next-token prediction as iterative energy descent over candidate continuations, exchanging the rigid autoregressive factorization for a more flexible test-time inference procedure. Third, LeCun and collaborators have developed the JEPA family~\cite{lecun2022,i-jepa,v-jepa} as an EBM-flavored alternative to autoregressive generative modeling for image and video. New theoretical instruments~\cite{ebm_neurips2025} continue to fill in the picture.

Against this backdrop the DBM~\cite{dbm} --- the classical deep EBM, predating the contemporary revival by more than a decade --- has not received commensurate attention as an applied tool. Its compatibility with partial observation is native rather than retrofitted, and that is the property this paper exploits.

\subsection{Boltzmann machines and the DBM}\label{sec:bg-bm}

A Boltzmann machine is a stochastic recurrent network of binary units with symmetric interactions and an energy $E(s) = -s^\top W s / 2 - b^\top s$, where the configuration $s$ comprises both visible and hidden units~\cite{bm_training}. The restricted Boltzmann machine (RBM)~\cite{smolensky_rbm,hinton_rbm} removes within-layer connections, factoring the energy into bipartite blocks and admitting a closed-form free energy. The DBM~\cite{dbm} stacks two or more RBM-style layers, restoring deep-network expressivity while retaining tractable mean-field inference: a
factorized variational posterior over the hidden layers can be optimized by coordinate ascent, each update reducing to a sigmoid of the weighted activations from neighboring layers. Training combines greedy layer-wise RBM pre-training with joint persistent contrastive divergence on the full model, with the partition function estimated by annealed importance sampling~\cite{salakhutdinov_murray_2008} when a likelihood bound is needed.

\paragraph{Relation to factor analysis.}
A useful conceptual anchor is that the DBM can be read as a non-linear, multilayer extension of factor analysis. A single RBM with linear hidden units and Gaussian observations reduces to classical factor analysis with mean-field inference; replacing the linear hidden activation with a sigmoid --- and the observations with Bernoulli units, as here --- yields a non-linear single-layer factor model, and stacking these layers yields the multilayer non-linear case~\cite{kamakura_wedel_1997,dbm}. The mean-field updates (\S\ref{sec:method-unroll}) are the direct analogue of expectation maximization in factor analysis. The factor-analytic baseline (\S\ref{sec:res-baselines}) is therefore a comparison against the DBM's own linear single-layer ancestor.

\subsection{Statistical data fusion}

Statistical data fusion has been approached with a variety of methodologies, including propensity-score matching~\cite{propensityscore}, clustering-based methods~\cite{clustering_fusion}, and factor-analytic models~\cite{kamakura_wedel_1997}. In the broader setting of missing-value imputation, common choices include multivariate imputation by chained equations (MICE)~\cite{mice}, random-forest imputation~\cite{missforest}, and deep generative models~\cite{vae,miwae}. We compare against representatives of all of these families (\S\ref{sec:res-baselines}).

Despite their natural fit, DBMs have seen little use in data fusion. An earlier preliminary proposal~\cite{niimi_hoshino_2017} suggested casting the problem as a DBM with per-sample observation masks --- missing visible dimensions are marginalized during training, and mean-field inference fills them in at test time --- but did not provide a formal derivation or numerical evaluation. We adopt that generative criterion as the first of two training stages (\S\ref{sec:method-mll}) and show that on its own it is the weakest arm we test; the contribution of this paper is what is built on top of it.

The closest analogue outside marketing is multimodal learning. The multimodal DBM~\cite{multimodal_dbm} joins modality-specific pathways, such as images and text tags, through a shared hidden layer, and fills in an absent modality by inference conditioned on the one that is present; this is the same operation as our cross-block conditioning, with modalities in place of outcome blocks. What differs is where the association is learned from. A multimodal model sees its modalities together in training, whereas at $p_\mathrm{complete}=0$ no row observes $Y_A$ and $Y_B$ together, and their association is identified only through the common block $X$ (\S\ref{sec:method-instantiation}). Data fusion is, in this sense, the multimodal problem with the paired examples removed.

\subsection{Multi-prediction training and the fusion regime}
\label{sec:bg-mpdbm}

The Multi-Prediction DBM (MP-DBM)~\cite{mp-dbm} was proposed specifically for missing-value prediction with DBMs. It replaces greedy layer-wise pre-training with a multi-prediction objective: at each minibatch a stochastic mask partitions the visible vector into ``observed'' and ``predicted'' subsets, and the model is trained with a cross-entropy loss on the predicted set, with the gradient flowing through the unrolled mean-field inference. The design principle --- train the parameters for the inference procedure that will be used at test time --- is the one we retain.

The objective requires ground-truth values for the predicted subset, and therefore presupposes fully observed training samples. This is precisely what the fusion setting withholds: source-$A$ rows lack $Y_B$ entirely and source-$B$ rows lack $Y_A$ entirely, so a random subset of visible dimensions
will in general contain entries no row can score. MP-DBM is thus undefined at $p_\mathrm{complete}=0$, the traditional data-fusion regime of Kamakura and Wedel~\cite{kamakura_wedel_1997}. \S\ref{sec:method-obmp} shows that the restriction needed to repair this is mild --- draw the targets from the observed set instead of from all visible dimensions --- and that the repaired criterion contains MP-DBM as the special case in which every row is complete.

\subsection{Missing-data taxonomy and our regime}

Rubin's taxonomy~\cite{rubin1976} distinguishes three regimes by the relationship between the missingness mechanism and the underlying values: missing completely at random (MCAR, independent of all values), missing at random (MAR, independent of the missing values given the observed ones), and missing not at random (MNAR, where missingness can depend on the unobserved values themselves).

The data-fusion setting is MAR conditional on the source indicator $z \in \{A, B\}$, which is part of the design rather than of nature: by construction source-$A$ rows never observe $Y_B$ and source-$B$ rows never observe $Y_A$, irrespective of what those values would have been. The likelihood-based MAR-correct response is to marginalize the missing dimensions out of the training objective, which is what \eqref{eq:Lo} does, and the identifiability argument for OBMP (\S\ref{sec:method-instantiation}) rests on the same conditional independence. MNAR regimes are out of scope here.

\section{Methodology}\label{sec:methodology}

We formalize a fine-tuning criterion for the Deep Boltzmann Machine that draws a discriminative training signal from partially observed rows. The criterion, which we call \emph{observed-block multi-prediction} (OBMP), restricts the multi-prediction objective of the MP-DBM~\cite{mp-dbm} to entries that are genuinely observed under a per-sample mask. It is therefore well-defined in the traditional data-fusion regime in which \emph{no} training row is fully observed, where the original criterion is not.

\subsection{DBM with per-sample observation masks}\label{sec:method-setup}

Consider a DBM with one visible layer $v \in \{0,1\}^{D_v}$ and $L$ hidden layers $h^{(1)}, \ldots, h^{(L)}$, with $h^{(l)} \in \{0,1\}^{D_{h_l}}$. Under the standard Bernoulli--Bernoulli parametrization~\cite{dbm}, the joint energy is
\begin{align}\label{eq:dbm-energy}
E(v, h^{(1{:}L)}; \theta) = - a^\top v - v^\top W^{(0)} h^{(1)} - \sum_{l=1}^{L-1} h^{(l)\top} W^{(l)} h^{(l+1)} - \sum_{l=1}^{L} b^{(l)\top} h^{(l)},
\end{align}
with parameters $\theta = \{a, b^{(1{:}L)}, W^{(0{:}L-1)}\}$ and $p_\theta(v, h^{(1{:}L)}) = Z(\theta)^{-1}\exp(-E(v, h^{(1{:}L)}; \theta))$.

Each training sample $i$ carries an observation mask $m_i \in \{0,1\}^{D_v}$, with $m_{ij}=1$ iff visible dimension $j$ is observed in sample $i$. In data fusion the visible dimensions partition into three blocks $\{X, Y_A, Y_B\}$: a common covariate block $X$ observed in every row, and two outcome blocks observed in disjoint panels. The uppercase names denote sets of visible dimensions; we write the matching lowercase symbols for the subvectors of $v$ on those dimensions, $v = (x, y_A, y_B)$ with $x \in \{0,1\}^{k}$, $y_A \in \{0,1\}^{d_a}$ and $y_B \in \{0,1\}^{d_b}$, so that $D_v = k + d_a + d_b$. Writing $\mathbf{1}_B$ for the indicator vector of a block $B$, each row falls into one of three groups,
\begin{align}\label{eq:masks}
m_i =
\begin{cases}
\mathbf{1}                                & \text{complete},\\
\mathbf{1}_X + \mathbf{1}_{Y_A}           & \text{source } A,\\
\mathbf{1}_X + \mathbf{1}_{Y_B}           & \text{source } B,
\end{cases}
\end{align}
and we write $p_\mathrm{complete}$ for the fraction of complete rows and $n_{\mathrm{complete}}$ for their count. The regime of interest throughout this paper is $p_\mathrm{complete}=0$: the two panels never overlap, so $Y_A$ and $Y_B$ are never observed together in training.

\subsection{Stage one: generative pre-training}\label{sec:method-mll}

The mask-aware marginal log-likelihood
\begin{align}\label{eq:Lo}
\mathcal{L}^o(\theta)
 = \sum_i \log p_\theta(v^{\mathrm{obs}}_i) 
 = \sum_i \log \!\!\sum_{v^{\mathrm{miss}}_i, h_i}\!\! \exp\bigl(-E(v_i, h_i; \theta)\bigr)
   - n_{\mathrm{train}} \log Z(\theta)
\end{align}
remains well-defined for any observation pattern, since missing visible dimensions are folded into the latent variables and summed out. We approximate the positive phase by a factorized mean-field posterior and the negative phase by persistent contrastive divergence over the full DBM state, giving the gradient
\begin{align}\label{eq:Lo-grad}
\nabla_\theta \mathcal{L}^o(\theta)
 \approx - \sum_i \mathbb{E}_{q_i}\!\bigl[\nabla_\theta E(v_i, h_i; \theta)\bigr] 
   + n_{\mathrm{train}}\, \mathbb{E}_{\tilde p_\theta}\!\bigl[\nabla_\theta E(v, h; \theta)\bigr].
\end{align}
We refer to a DBM trained by \eqref{eq:Lo} as an ML-DBM and use it as the initialization for stage two. It is a purely generative criterion: no term in \eqref{eq:Lo} distinguishes covariates from outcomes.

\subsection{Why multi-prediction training is undefined here}\label{sec:method-gap}

The MP-DBM~\cite{mp-dbm} replaces \eqref{eq:Lo} with a discriminative criterion. At each minibatch it draws a random subset $S$ of visible dimensions, runs mean-field inference conditioned on $v_{\bar S}$, and penalises the cross-entropy between the inferred marginals on $S$ and the true values $v_S$. The construction presupposes that $v_S$ is available, i.e. that the row is fully observed; on a row from source $A$, any $S$ intersecting $Y_B$ has no ground truth to compare against. Consequently the criterion draws gradient signal only from the $n_{\mathrm{complete}}$ complete rows, and is undefined outright when $n_{\mathrm{complete}}=0$.

\subsection{Observed-block multi-prediction}\label{sec:method-obmp}

OBMP removes this restriction by choosing the prediction targets \emph{inside} the observed set. For each row we introduce two binary masks, a \emph{conditioning mask} $c_i \in \{0,1\}^{D_v}$ marking the entries supplied to inference, and a \emph{target mask} $t_i \in \{0,1\}^{D_v}$ marking the entries whose known values define the loss, subject to
\begin{equation}\label{eq:mask-constraints}
c_i \odot t_i = \mathbf{0},
\qquad
t_i \preceq m_i,
\end{equation}
where $\odot$ is the elementwise product and $\preceq$ holds elementwise. The first constraint keeps a target from being fed to the model as evidence; the second is the substantive one, and states that every target must be a genuinely observed value. Entries that are neither conditioned on nor targeted are treated as missing and marginalized, exactly as in \eqref{eq:Lo}.

Let $\mu^{(v)}(c_i) \in [0,1]^{D_v}$ denote the visible marginals returned by the mean-field procedure (\S\ref{sec:method-unroll}) when the entries selected by $c_i$ are clamped to their observed values, suppressing its dependence on $v_i$ and $\theta$ in the notation. With $\ell(p, y) = -y \log p - (1-y)\log(1-p)$ the elementwise binary cross-entropy, the OBMP criterion is
\begin{equation}\label{eq:obmp}
\mathcal{L}^{\mathrm{OBMP}}(\theta)
 = \frac{1}{|\mathcal{I}|} \sum_{i \in \mathcal{I}}
   \frac{\sum_{j} t_{ij}\,
         \ell\bigl(\mu^{(v)}_{ij}(c_i),\, v_{ij}\bigr)}
        {\sum_{j} t_{ij}},
\end{equation}
where $\mathcal{I} = \{ i : \sum_j t_{ij} > 0 \}$. The inner normalization by $\sum_j t_{ij}$ weights every row equally regardless of how many targets it contributes, so that rows from a panel with a wider outcome block do not dominate the gradient; $\mathcal{I}$ excludes rows with no target at all. In practice $\mu^{(v)}$ is clamped to $[\varepsilon, 1-\varepsilon]$ before the logarithm.

Two properties follow directly from \eqref{eq:mask-constraints}. First, OBMP is defined for \emph{any} mask pattern, including $m_i \neq \mathbf{1}$ for every $i$: the targets are drawn from what each row actually observes, so every row contributes gradient. Second, when every row is complete ($m_i \equiv \mathbf{1}$) and $c_i = \mathbf{1} - t_i$ with $t_i$ drawn at random, \eqref{eq:obmp} reduces to the MP-DBM criterion. OBMP is thus a strict generalization of multi-prediction training to arbitrary observation patterns, and inherits its interpretation as training the model to be a good inference machine rather than a good density model.

In the general case the two masks are obtained by splitting the observed set at random: draw a per-row keep probability $\pi_i \sim \mathcal{U}(\pi_{\min}, \pi_{\max})$, keep each observed entry independently with probability $\pi_i$ to form $c_i$, and target the rest, 
\begin{equation}\label{eq:random-split}
c_i = m_i \odot \rho_i,
\qquad
t_i = m_i - c_i,
\qquad
\rho_{ij} \sim \mathrm{Bern}(\pi_i).
\end{equation}
This recovers the stochastic masking of MP-DBM restricted to the observed support. The data-fusion instantiation below replaces it with a deterministic split that matches the deployment pattern.

\subsection{Differentiable mean-field unroll}\label{sec:method-unroll}

Evaluating \eqref{eq:obmp} requires $\mu^{(v)}$ to be a differentiable
function of $\theta$. We therefore run mean-field inference as a fixed-length
unrolled computation and backpropagate through all of it, rather than treating
the fixed point as a constant. Write $\mu^{(v),\tau}$ and $\mu^{(l),\tau}$ for
the visible and hidden marginals after $\tau$ passes, and let $\sigma$ denote
the elementwise logistic function. The initialization clamps the conditioned
entries and sets the remainder from the visible bias, then propagates upward:
\begin{align}
\mu^{(v),0} &= c \odot v + (\mathbf{1}-c) \odot \sigma(a), \label{eq:mf-init-v}\\
\mu^{(l),0} &= \sigma\bigl(\mu^{(l-1),0} W^{(l-1)} + b^{(l)}\bigr),
   \quad l = 1, \ldots, L, \label{eq:mf-init-h}
\end{align}
with the convention $\mu^{(0),\tau} \coloneqq \mu^{(v),\tau}$. Each subsequent
pass $\tau = 1, \ldots, T$ updates the visible marginals from the first hidden
layer, re-clamping the conditioned entries, and then sweeps the hidden layers
bottom-up:
\begin{align}
\mu^{(v),\tau} &= c \odot v + (\mathbf{1}-c) \odot \sigma\bigl(\mu^{(1),\tau-1} W^{(0)\top} + a\bigr), \label{eq:mf-v}\\
\mu^{(l),\tau} &= \sigma\bigl(\mu^{(l-1),\tau} W^{(l-1)} + b^{(l)} + \mu^{(l+1),\tau-1} W^{(l)\top}\bigr), \label{eq:mf-h} \\
\mu^{(L),\tau} &= \sigma\bigl(\mu^{(L-1),\tau} W^{(L-1)} + b^{(L)}\bigr), \label{eq:mf-htop}
\end{align}
where $l = 1, \ldots, L-1$. Equation~\eqref{eq:mf-h} carries the top-down term at pass $\tau-1$ because
the sweep proceeds bottom-up within a pass, so the layer above has not yet
been refreshed. Conditioned entries enter \eqref{eq:mf-h} through
$\mu^{(v),\tau}$ but are never themselves updated, so the clamping in
\eqref{eq:mf-v} is exact at every pass. Optionally each update may be damped,
$\mu \leftarrow (1-\lambda)\mu^{\mathrm{new}} + \lambda \mu^{\mathrm{old}}$;
we use $\lambda = 0$ throughout.

We take $\mu^{(v)} \coloneqq \mu^{(v),T}$ in \eqref{eq:obmp}. Because
\eqref{eq:mf-init-v}--\eqref{eq:mf-htop} are compositions of affine maps and
logistic nonlinearities, the whole unroll is differentiable, and the gradient
of \eqref{eq:obmp} flows through all $T$ passes. The criterion therefore
trains the parameters \emph{for the inference procedure that will be used at
test time}, which is the same design principle that motivates MP-DBM; the
difference is only in which entries are allowed to serve as targets.

\subsection{Instantiation for data fusion}\label{sec:method-instantiation}

In the fusion setting the split is deterministic. We condition on the common
block and target every observed outcome:
\begin{equation}\label{eq:fusion-split}
c_i = \mathbf{1}_X,
\qquad
t_i = m_i - \mathbf{1}_X = m_i \odot (\mathbf{1}_{Y_A} + \mathbf{1}_{Y_B}),
\end{equation}
which by \eqref{eq:masks} gives $t_i = \mathbf{1}_{Y_A}$ for a source-$A$ row,
$t_i = \mathbf{1}_{Y_B}$ for a source-$B$ row, and
$t_i = \mathbf{1}_{Y_A} + \mathbf{1}_{Y_B}$ for a complete row. The
constraints \eqref{eq:mask-constraints} hold by construction, and
$\mathcal{I}$ contains every row, so all $n_{\mathrm{train}}$ rows contribute gradient at
$p_\mathrm{complete}=0$.

This choice makes the criterion identifiable under the missingness regime of data fusion. Missingness is determined by the source indicator, which is independent of the outcome values given $X$; conditioning on $X$ alone and predicting the observed outcome block therefore targets $p(y_A \mid x)$ on source-$A$ rows and $p(y_B \mid x)$ on source-$B$ rows, both of which are identified from the observed data. No term in \eqref{eq:obmp} requires the joint $p(y_A, y_B \mid x)$, which is not identified without complete rows.

The inference (\S\ref{sec:method-conditioning}) is a different matter, and we state the asymmetry plainly because it bounds what the experiments can show. Predicting $y_A$ from $x$ \emph{and} $y_B$ requires $p(y_A \mid x, y_B)$, a functional of exactly the joint the objective avoided, and nothing in fusion data identifies it. What the model supplies in its place is an inductive bias: the hidden layers carry all dependence between the outcome blocks, so conditioning on $y_B$ acts through them. This is the same structure as the conditional independence assumption underlying factor-analytic and matching approaches to statistical matching~\cite{kamakura_wedel_1997,raessler2004}, where $Y_A$ and $Y_B$ are taken to be independent given the common variables; here they are taken to be independent given the common variables \emph{and} the hidden state. That is a weaker assumption, since the hidden state is learned rather than fixed to $X$, but it is an assumption and it is not testable on fusion data. Our test rows happen to observe both blocks, so we can check after the fact whether it paid off; a practitioner with genuine fusion data cannot.

\subsection{Inference-time conditioning}\label{sec:method-conditioning}

The conditioning mask is a property of the inference call, not of the trained
parameters, so a single OBMP checkpoint can be evaluated under different
conditioning sets. We distinguish two.

\paragraph{Cross-block conditioning.} To predict outcome block $Y_A$ we clamp
everything else that the row observes, including the other outcome block:
\begin{equation}\label{eq:cond-cross}
c = \mathbf{1} - \mathbf{1}_{Y_A}
  = \mathbf{1}_X + \mathbf{1}_{Y_B},
\end{equation}
and symmetrically $c = \mathbf{1}_X + \mathbf{1}_{Y_A}$ to predict $Y_B$. This
matches the operational task in panel fusion: a source-$A$ row is one whose
$Y_A$ is recorded and whose $Y_B$ must be filled in, so $Y_A$ is available as
evidence precisely when $Y_B$ is the quantity of interest.

\paragraph{$X$-only conditioning.} Alternatively both outcome blocks are
hidden and predicted in a single pass,
\begin{equation}\label{eq:cond-xonly}
c = \mathbf{1}_X,
\end{equation}
which is the conditioning set used during training \eqref{eq:fusion-split},
and the only one available to a discriminative model fitted on $X$ alone.

The gap between \eqref{eq:cond-cross} and \eqref{eq:cond-xonly}, measured on a
single checkpoint, isolates the value of conditioning on one outcome block
when predicting the other. It is a quantity that no model trained to map $X$
to outcomes can produce, and, because $Y_A$ and $Y_B$ are never observed
together in training, it is supplied entirely by the generative structure of
the joint model rather than by a fitted association between the two blocks.

\subsection{Two-stage procedure}\label{sec:method-procedure}

Training proceeds in two stages on the same architecture and the same rows.
Stage one fits the ML-DBM by \eqref{eq:Lo-grad}, initialising the visible bias
from the observed marginals $a_j = \mathrm{logit}(\bar v_j)$ computed over the
entries with $m_{ij}=1$, since with $p_\mathrm{complete}=0$ there is no
complete subset on which to pre-train layerwise. Stage two initialises from
the stage-one parameters and minimises \eqref{eq:obmp} with the fusion split
\eqref{eq:fusion-split}, selecting the checkpoint by validation performance.
Stage two changes the criterion but not the model: the parameters
$\theta$ remain those of the DBM in \eqref{eq:dbm-energy}, and the same
mean-field procedure is used for training and for both inference modes of
\S\ref{sec:method-conditioning}.

\section{Experimental design}\label{sec:exp}

\subsection{Datasets and fusion protocol}\label{sec:exp-data}

We report two datasets, chosen to differ in kind: a behavioral purchase panel and survey microdata. Blocks are assigned by subject matter, not at random, so that each panel is one a real study might have collected.

\textbf{Instacart} is an online-grocery panel binarized to purchase indicators at the aisle level, with aisles assigned to blocks by department. $X$ holds fresh-food departments (produce, dairy and eggs, beverages, bakery, meat and seafood), $Y_A$ the ambient-food departments (pantry, frozen, snacks, breakfast, canned goods, dry goods, deli, $d_a = 53$) and $Y_B$ the non-food departments (personal care, household, babies, pets, alcohol, international, bulk, other, $d_b = 46$), so $D_v = k + 99$. Restricting $X$ to width $k$ keeps its $k$ most-purchased aisles and drops the rest from the visible layer entirely; they are not moved into an outcome block. We draw from a pool of $80{,}000$ users and hold out $n_{\mathrm{val}} = n_{\mathrm{test}} = 5{,}000$.

\textbf{ACS-housing} is built from the 2024 American Community Survey 1-year Public Use Microdata Sample (PUMS) for California, person and housing records, released by the US Census Bureau.\footnote{\url{https://www.census.gov/programs-surveys/acs/microdata.html}} We keep one row per occupied housing unit: the householder, aged 18 or over, joined to the housing record, with group quarters excluded. One row per household keeps housemates, who share every housing variable, from straddling the train/test split. The result has $141{,}179$ rows; survey weights are not used. $X$ holds demographics (age bands, sex, race, Hispanic origin, marital status, educational attainment, citizenship, language spoken at home, limited English, veteran status, household size and presence of children, $30$ dimensions in all), $Y_A$ labor and income (employment status, class of worker, usual weekly hours, weeks worked, wage tertiles, personal-income quartiles, and receipt of self-employment, retirement, Social Security and public-assistance income, $d_a = 22$) and $Y_B$ housing (tenure, household-income quartiles, SNAP receipt, vehicles available, structure type, gross rent of at least $30$\% of income, and no internet access, $d_b = 17$), so $D_v = k + 39$. All variables are one-hot encoded or quantile-binned. As on Instacart, restricting $X$ to width $k$ keeps its $k$ most common columns. We draw from a pool of $80{,}000$ training rows and hold out $n_{\mathrm{val}} = n_{\mathrm{test}} = 5{,}000$. The fusion task is the realistic one of joining a labor-force survey to a housing survey that share demographics.

The two datasets also differ in block geometry, with outcome blocks of $53$ and $46$ dimensions on Instacart and of $22$ and $17$ on ACS-housing, and in how much headroom each block leaves (\S\ref{sec:res-blocks}). ACS-housing is a US federal work in the public domain, so the derived matrices are released on Hugging Face,\footnote{\url{https://huggingface.co/datasets/jniimi/datafusion-acs}} with the split of our first seed arranged so that its first $n_{\mathrm{train}}$ training rows are the training set we use at that size with all $30$ covariates; the Instacart matrices cannot be redistributed under that dataset's terms.

Every training row is assigned to source $A$ or source $B$ with probability $p_a = 0.5$ and masked according to \eqref{eq:masks}. We set $p_\mathrm{complete}=0$ throughout: \emph{no} training row observes both outcome blocks. This is the regime in which the MP-DBM criterion is undefined, and it is the regime that motivates OBMP. Validation and test rows retain both blocks, since scoring requires ground truth for the block being predicted; the conditioning sets (\S\ref{sec:method-conditioning}) control what the model is allowed to see.

\subsection{The $(n_{\mathrm{train}},k)$ grids}\label{sec:exp-grid}

We vary two quantities jointly: the number of training rows $n_{\mathrm{train}} \in \{500, 1000, 2000, 5000, 10{,}000\}$, drawn from the pool described above and disjoint from the $n_{\mathrm{val}}$ validation and $n_{\mathrm{test}}$ test rows, and the width of the common block, $k \in \{5, 10, 20, 35\}$ on Instacart and $k \in \{5, 10, 20, 30\}$ on ACS-housing, whose $X$ block has $30$ dimensions in total. Each cell is repeated over five seeds, for $100$ runs per dataset. The two axes are the natural stress directions for a data-fusion method: $n_{\mathrm{train}}$ controls how much evidence is available, and $k$ controls how much of the outcome variation the shared covariates can explain. Training subsets are nested within a seed, drawn by a fixed permutation rule so that a larger $n_{\mathrm{train}}$ contains the rows of every smaller $n_{\mathrm{train}}$.

\subsection{Methods compared}\label{sec:exp-methods}

Five arms share the DBM architecture --- $L = 2$ hidden layers with $D_{h_1} = 64$ and $D_{h_2} = 32$ --- and are run on every cell of both grids:

\begin{itemize}[leftmargin=*,topsep=2pt,itemsep=1pt]
\item \textbf{ML-DBM}: generative training by \eqref{eq:Lo-grad} alone,
      no discriminative fine-tuning.
\item \textbf{OBMP (cross-block)}: the ML-DBM fine-tuned by \eqref{eq:obmp},
      evaluated with the conditioning of \eqref{eq:cond-cross}.
\item \textbf{OBMP ($X$-only)}: \emph{the same checkpoint} evaluated with
      \eqref{eq:cond-xonly}. The difference between these two arms is measured
      on identical parameters and so is free of training confounds.
\item \textbf{MLP}: a randomly initialized sigmoid network
      $k \to 64 \to 32 \to (d_a{+}d_b)$ trained on the same observed-block
      cross-entropy, with weight decay chosen on validation. It matches the
      DBM's capacity and objective but has no generative pre-training and no
      cross-block inference path.
\item \textbf{$X$-logistic}: per-dimension logistic regression on $X$, with the
      regularization strength chosen on validation.
\end{itemize}

The MLP and $X$-logistic arms are chosen to bracket the two mechanisms that could explain a DBM advantage without invoking the joint model: a nonlinear shared representation, and a well-tuned conditional mean. We additionally compare against baselines spanning marginal predictors, nearest neighbours, chained equations, classical data fusion, factor analysis and deep generative imputation (Table~\ref{tab:baselines}), each tuned on validation over a grid given in Appendix~\ref{app:repro}. The four joint imputers among them (MICE, factor analysis, MissForest and MIWAE) can take the other outcome block as evidence, so each is evaluated under both conditioning sets of \S\ref{sec:method-conditioning}. All selection uses $X$-only validation, including for the cross-block evaluations, because fusion data contain no row observing both blocks on which cross-block conditioning could be validated; this is the rule OBMP itself follows (\S\ref{sec:exp-budget}). In the summary table (Table~\ref{tab:main-accuracy}) each joint imputer is credited with whichever conditioning set scores better on that cell's test split, a choice no practitioner could make, and one that favours the baselines.

\paragraph{Sample-size parity.} Every method, including every baseline, is fitted on the same $n_{\mathrm{train}}$ training rows, verified per cell by recomputing the tuned $X$-logistic arm inside the baseline runner and matching it against the value recorded by the main run (deviation exactly zero on all $200$ runs of the two datasets, and again on all $200$ runs of the tuned baselines).

\subsection{Training budget and checkpoint selection}\label{sec:exp-budget}

OBMP fine-tuning and the MLP arm are trained for up to 200 epochs on every cell, evaluated on validation every five epochs with checkpoint selection and a patience of 60. Holding the two arms to one budget is what makes the decomposition (\S\ref{sec:res-decomp}) interpretable: a difference between them is a difference of method, not of how long each was allowed to train. The budget is generous enough that OBMP selects its final epoch strictly below the cap in 98 of the 100 runs on each dataset. The mean-field unroll (\S\ref{sec:method-unroll}) runs a fixed $T = 10$ passes during training, since the loss has to differentiate through all of them. At evaluation, where no gradient is needed, $T$ is a cap: the sweep stops early once no coordinate moves by more than $10^{-4}$, and we set the cap to $15$. \S\ref{sec:res-depth} reports what happens when it is varied.

Both arms select their checkpoint on \emph{$X$-only} validation accuracy, the conditioning that the training objective itself uses \eqref{eq:fusion-split}. This matters for the contrast (\S\ref{sec:method-conditioning}). Selecting the OBMP checkpoint on cross-block validation instead --- the conditioning under which it will be reported --- and then subtracting its $X$-only accuracy would let the selection contribute to the gap: with roughly forty candidate epochs per cell, choosing the one that scores best under cross-block evaluation picks up the upper tail of the validation noise in exactly the direction the contrast is measured. We quantify that effect (\S\ref{sec:res-decomp}) by tracking both checkpoints from the same run.

\subsection{Metrics}\label{sec:exp-metrics}

All metrics are computed over the positions being predicted, never over observed ones. We report a dimension-weighted combination of the two outcome blocks,
\begin{equation}\label{eq:combined}
\mathrm{combined} = \frac{d_a \cdot s_{Y_A} + d_b \cdot s_{Y_B}}{d_a + d_b},
\end{equation}
where $s_B$ is the metric on block $B$: either accuracy at a threshold of $0.5$ or binary cross-entropy. Accuracy is the headline number throughout. We quote cross-entropy only where calibration is the point (\S\ref{sec:disc-size}); the three baselines that emit hard $0/1$ values rather than probabilities are not comparable on it at all.

Every test in this paper is paired and clustered on seed. Runs sharing a seed share a test split and nested training subsets, so they are not independent replications; averaging within seed and testing the five means gives four degrees of freedom rather than the spuriously large ones a run-level test would assume. The seed-to-seed standard deviation of the contrasts we report is $0.012$ to $0.048$\,pp, so the conclusions survive the stricter accounting, but the $p$-values are correspondingly less extreme.

\section{Results}\label{sec:results}

\subsection{OBMP against every method}\label{sec:res-main}

Table~\ref{tab:main-accuracy} reports both grids, with the strongest tuned baseline of each cell alongside the five arms. OBMP with cross-block conditioning is the best method in $18$ of the $20$ Instacart cells and $19$ of the $20$ ACS-housing cells, $37$ of $40$ in all. The three exceptions are all at $k{=}5$, the narrowest common block, and all are small: on Instacart at $n_{\mathrm{train}}{=}500$ and $n_{\mathrm{train}}{=}2000$, MIWAE with cross-block conditioning is ahead by $0.006$ and $0.035$\,pp, and on ACS-housing at $n_{\mathrm{train}}{=}1000$ the MLP is ahead by $0.005$\,pp. The margin over tuned $X$-logistic narrows as $n_{\mathrm{train}}$ grows on Instacart and widens on ACS-housing (\S\ref{sec:res-trends}).

Two features are worth noting before any decomposition. First, ML-DBM alone is the weakest arm on both datasets, and on ACS-housing it is degenerate: it equals the column-mean baseline exactly in $99$ of $100$ runs and trails tuned $X$-logistic by $0.17$ to $3.3$\,pp. On Instacart it moves a little with $n_{\mathrm{train}}$ and $k$ but still trails tuned $X$-logistic by $1.4$ to $2.5$\,pp, and its $Y_B$ accuracy never leaves the base rate. Whatever OBMP contributes on either dataset, it is not inherited from the generative stage. Second, the ranking of the two OBMP arms is stable: cross-block conditioning beats $X$-only conditioning in every cell of both grids, on parameters that are identical.

\begin{table}[htbp]
\centering
\caption{Test combined accuracy [\%] on the $(n_{\mathrm{train}},k)$ grids, mean over five seeds (standard deviation in parentheses). Best per cell in bold. \emph{best BL} is the strongest of the other baselines on that cell (\S\ref{sec:exp-methods}), tuned on validation and fitted on the same training rows, with each joint imputer credited with whichever conditioning set scores better on that cell. The three ablations each remove one component of OBMP and share its training budget (\S\ref{sec:exp-budget}).}
\label{tab:main-accuracy}
\begin{tabular}{rr c cc ccc}
\toprule
 & & proposed & \multicolumn{2}{c}{baselines} & \multicolumn{3}{c}{ablations of OBMP} \\
\cmidrule(lr){3-3} \cmidrule(lr){4-5} \cmidrule(lr){6-8}
$n_{\mathrm{train}}$ & $k$ & OBMP & $X$-log. & best BL & ML-DBM & $X$-only & MLP \\
\midrule
\multicolumn{8}{l}{\emph{Instacart}} \\
500 & 5 & 83.24 (0.08) & 82.93 (0.05) & \textbf{83.25} & 81.50 (0.07) & 83.06 (0.11) & 82.68 (0.56) \\
 & 10 & \textbf{83.50 (0.12)} & 82.94 (0.07) & 83.28 & 81.50 (0.03) & 83.30 (0.08) & 83.17 (0.10) \\
 & 20 & \textbf{83.80 (0.07)} & 83.35 (0.06) & 83.79 & 81.52 (0.14) & 83.66 (0.08) & 83.50 (0.09) \\
 & 35 & \textbf{83.86 (0.12)} & 83.38 (0.09) & 83.86 & 81.51 (0.08) & 83.70 (0.08) & 83.60 (0.07) \\
\addlinespace
1000 & 5 & \textbf{83.32 (0.08)} & 83.05 (0.07) & 83.24 & 81.54 (0.04) & 83.11 (0.09) & 83.06 (0.07) \\
 & 10 & \textbf{83.64 (0.08)} & 83.21 (0.09) & 83.39 & 81.56 (0.06) & 83.37 (0.07) & 83.33 (0.07) \\
 & 20 & \textbf{84.03 (0.04)} & 83.63 (0.04) & 83.86 & 81.55 (0.09) & 83.86 (0.04) & 83.80 (0.05) \\
 & 35 & \textbf{84.07 (0.06)} & 83.71 (0.07) & 83.99 & 81.59 (0.06) & 83.94 (0.06) & 83.91 (0.07) \\
\addlinespace
2000 & 5 & 83.27 (0.10) & 83.11 (0.09) & \textbf{83.31} & 81.57 (0.07) & 83.14 (0.08) & 83.13 (0.07) \\
 & 10 & \textbf{83.71 (0.06)} & 83.39 (0.07) & 83.35 & 81.57 (0.09) & 83.42 (0.07) & 83.44 (0.08) \\
 & 20 & \textbf{84.14 (0.08)} & 83.90 (0.07) & 83.95 & 81.58 (0.04) & 83.97 (0.07) & 83.95 (0.06) \\
 & 35 & \textbf{84.20 (0.08)} & 83.95 (0.03) & 84.07 & 81.61 (0.08) & 84.07 (0.05) & 84.08 (0.02) \\
\addlinespace
5000 & 5 & \textbf{83.29 (0.12)} & 83.15 (0.07) & 83.21 & 81.62 (0.10) & 83.17 (0.08) & 83.17 (0.08) \\
 & 10 & \textbf{83.74 (0.04)} & 83.48 (0.07) & 83.36 & 81.59 (0.05) & 83.47 (0.06) & 83.48 (0.06) \\
 & 20 & \textbf{84.26 (0.04)} & 84.07 (0.05) & 83.98 & 81.64 (0.07) & 84.06 (0.05) & 84.05 (0.05) \\
 & 35 & \textbf{84.40 (0.04)} & 84.16 (0.03) & 84.14 & 81.70 (0.05) & 84.22 (0.02) & 84.23 (0.04) \\
\addlinespace
10000 & 5 & \textbf{83.27 (0.10)} & 83.16 (0.07) & 83.20 & 81.68 (0.09) & 83.18 (0.08) & 83.17 (0.07) \\
 & 10 & \textbf{83.79 (0.04)} & 83.50 (0.07) & 83.41 & 81.76 (0.09) & 83.49 (0.06) & 83.51 (0.05) \\
 & 20 & \textbf{84.32 (0.07)} & 84.12 (0.07) & 83.96 & 82.41 (0.12) & 84.12 (0.08) & 84.11 (0.07) \\
 & 35 & \textbf{84.47 (0.03)} & 84.26 (0.05) & 84.18 & 82.82 (0.14) & 84.26 (0.04) & 84.27 (0.04) \\
\midrule
\multicolumn{8}{l}{\emph{ACS-housing}} \\
500 & 5 & \textbf{80.30 (0.15)} & 80.26 (0.11) & 80.28 & 80.09 (0.04) & 80.24 (0.11) & 80.26 (0.13) \\
 & 10 & \textbf{81.13 (0.16)} & 80.79 (0.12) & 80.69 & 80.04 (0.13) & 80.76 (0.14) & 80.60 (0.16) \\
 & 20 & \textbf{82.44 (0.21)} & 81.95 (0.23) & 81.99 & 80.09 (0.04) & 82.14 (0.18) & 82.15 (0.13) \\
 & 30 & \textbf{82.64 (0.16)} & 82.13 (0.11) & 82.22 & 80.09 (0.04) & 82.23 (0.08) & 82.15 (0.11) \\
\addlinespace
1000 & 5 & 80.43 (0.14) & 80.35 (0.12) & 80.40 & 80.09 (0.04) & 80.39 (0.12) & \textbf{80.43 (0.07)} \\
 & 10 & \textbf{81.49 (0.13)} & 81.07 (0.08) & 80.79 & 80.09 (0.04) & 81.03 (0.09) & 81.11 (0.10) \\
 & 20 & \textbf{82.85 (0.11)} & 82.52 (0.08) & 82.17 & 80.09 (0.04) & 82.52 (0.07) & 82.50 (0.12) \\
 & 30 & \textbf{83.05 (0.12)} & 82.65 (0.11) & 82.44 & 80.09 (0.04) & 82.67 (0.13) & 82.71 (0.15) \\
\addlinespace
2000 & 5 & \textbf{80.59 (0.15)} & 80.40 (0.12) & 80.46 & 80.09 (0.04) & 80.51 (0.14) & 80.53 (0.06) \\
 & 10 & \textbf{81.83 (0.15)} & 81.20 (0.08) & 80.95 & 80.09 (0.04) & 81.34 (0.13) & 81.34 (0.11) \\
 & 20 & \textbf{83.15 (0.07)} & 82.80 (0.08) & 82.34 & 80.09 (0.04) & 82.78 (0.06) & 82.83 (0.08) \\
 & 30 & \textbf{83.56 (0.17)} & 83.04 (0.09) & 82.57 & 80.09 (0.04) & 83.10 (0.15) & 83.14 (0.13) \\
\addlinespace
5000 & 5 & \textbf{80.71 (0.13)} & 80.43 (0.08) & 80.45 & 80.09 (0.04) & 80.61 (0.05) & 80.60 (0.06) \\
 & 10 & \textbf{82.02 (0.13)} & 81.29 (0.06) & 81.19 & 80.09 (0.04) & 81.45 (0.11) & 81.51 (0.06) \\
 & 20 & \textbf{83.47 (0.09)} & 82.97 (0.08) & 82.59 & 80.09 (0.04) & 83.07 (0.11) & 83.12 (0.09) \\
 & 30 & \textbf{83.83 (0.18)} & 83.31 (0.08) & 82.72 & 80.09 (0.04) & 83.31 (0.13) & 83.39 (0.13) \\
\addlinespace
10000 & 5 & \textbf{80.85 (0.07)} & 80.43 (0.07) & 80.53 & 80.09 (0.04) & 80.67 (0.07) & 80.67 (0.07) \\
 & 10 & \textbf{82.12 (0.12)} & 81.32 (0.06) & 81.32 & 80.09 (0.04) & 81.52 (0.08) & 81.58 (0.06) \\
 & 20 & \textbf{83.66 (0.12)} & 83.03 (0.08) & 82.76 & 80.09 (0.04) & 83.17 (0.12) & 83.20 (0.09) \\
 & 30 & \textbf{83.99 (0.06)} & 83.41 (0.07) & 82.90 & 80.09 (0.04) & 83.47 (0.15) & 83.49 (0.11) \\
\bottomrule
\end{tabular}
\end{table}

\subsection{Against tuned baselines}\label{sec:res-baselines}

Table~\ref{tab:baselines} compares OBMP with every baseline, run by run. Each baseline is fitted on the same rows and tuned on validation where it has anything to tune; tuning alone raised the baselines by $0.2$ to $2.8$\,pp over their fixed settings, measured on a subset of the ACS-housing cells (Appendix~\ref{app:repro}). The joint imputers are scored twice, with both outcome blocks hidden and, on the indented rows, with the other block supplied as evidence, the setting in both cases chosen on $X$-only validation. OBMP is ahead of every baseline under either conditioning on both datasets, all at $p < 0.05$ clustered on seed. Against the methods that predict the missing block from $X$ alone the margins are tenths of a point: $+0.293$ and $+0.438$\,pp over $X$-logistic on Instacart and ACS-housing, $+0.413$ and $+0.701$ over $k$-NN, $+0.308$ and $+0.755$ over MICE, $+0.419$ and $+0.967$ over factor analysis, and $+0.254$ and $+1.100$ over MIWAE. The methods that return hard values (MissForest, hot deck and propensity matching) and the two constant predictors are several points behind.

\begin{table}[htbp]
\centering
\caption{OBMP against the baselines, all fitted on the same training rows and tuned on validation where they have anything to tune. The joint imputers are scored with both outcome blocks hidden and, on the indented rows, with the other block supplied, the setting still chosen on $X$-only validation as OBMP's is. $\Delta$ is the paired difference from OBMP [pp] and \emph{wins} counts runs where OBMP is ahead. $^{*}$ marks $p<0.05$ against zero, clustered on seed. $^{\dagger}$ returns hard $0/1$ values, so its cross-entropy is not comparable.}
\label{tab:baselines}
\begin{tabular}{l rr rr}
\toprule
 & \multicolumn{2}{c}{Instacart (100)} & \multicolumn{2}{c}{ACS-housing (100)} \\
\cmidrule(lr){2-3} \cmidrule(lr){4-5}
method & $\Delta$ & wins & $\Delta$ & wins \\
\midrule
$X$-logistic & $+0.293^{*}$ & 99/100 & $+0.438^{*}$ & 96/100 \\
$k$-NN & $+0.413^{*}$ & 100/100 & $+0.701^{*}$ & 96/100 \\
MICE & $+0.308^{*}$ & 97/100 & $+0.755^{*}$ & 96/100 \\
\quad with cross-block & $+2.337^{*}$ & 98/100 & $+5.382^{*}$ & 100/100 \\
factor analysis & $+0.419^{*}$ & 100/100 & $+0.967^{*}$ & 97/100 \\
\quad with cross-block & $+1.004^{*}$ & 99/100 & $+1.408^{*}$ & 99/100 \\
MIWAE & $+0.254^{*}$ & 95/100 & $+1.100^{*}$ & 99/100 \\
\quad with cross-block & $+0.227^{*}$ & 84/100 & $+1.953^{*}$ & 100/100 \\
MissForest$^{\dagger}$ & $+16.732^{*}$ & 100/100 & $+12.552^{*}$ & 100/100 \\
\quad with cross-block & $+32.810^{*}$ & 100/100 & $+19.603^{*}$ & 100/100 \\
hot deck$^{\dagger}$ & $+6.799^{*}$ & 100/100 & $+7.993^{*}$ & 100/100 \\
propensity matching$^{\dagger}$ & $+8.281^{*}$ & 100/100 & $+9.414^{*}$ & 100/100 \\
column mean & $+2.255^{*}$ & 100/100 & $+2.119^{*}$ & 100/100 \\
mode zero & $+2.493^{*}$ & 100/100 & $+2.326^{*}$ & 100/100 \\
\midrule
strongest per run & $+0.128^{*}$ & 80/100 & $+0.412^{*}$ & 93/100 \\
\bottomrule
\end{tabular}
\end{table}

The comparison that matters is with the strongest competitor on each run, taken over every baseline and, for the joint imputers, over both conditioning sets, all selected on $X$-only validation. OBMP is ahead of it by $+0.128$\,pp on Instacart, in $80$ of $100$ runs ($p = 0.002$, clustered on seed), and by $+0.412$\,pp on ACS-housing, in $93$ of $100$ ($p < 0.001$). Which competitor is strongest depends on the dataset. On Instacart it is MIWAE with cross-block conditioning in $8$ cells, MICE from $X$ alone in $7$ and MIWAE from $X$ alone in $5$; MIWAE with cross-block conditioning is also the closest single baseline there ($+0.227$\,pp, OBMP ahead in $84$ of $100$ runs). On ACS-housing it is $k$-NN in $11$ cells and MICE in $9$, and the same MIWAE falls further behind when it is conditioned on the other block ($+1.953$\,pp, $100$ of $100$) than when it is not.

Without the other block, OBMP has little to offer over tuned imputers. Restricting OBMP to its $X$-only arm and the baselines to $X$-only conditioning, OBMP is not ahead of the strongest competitor on Instacart ($-0.006$\,pp, $56$ of $100$ runs, $p = 0.29$) and only slightly ahead on ACS-housing ($+0.057$\,pp, $69$ of $100$). The $X$-only arm is a control in the decomposition, not the method we propose, and this comparison makes the decomposition's point from another direction: what separates OBMP from tuned imputers is not what it predicts from $X$, but what it does with the other outcome block.

\subsection{Conditioning blindly}\label{sec:res-conditioning}

In fusion no row observes both outcome blocks, so whether conditioning on the other block improves a given model cannot be checked on validation data; if it is used at all, it has to be applied blindly. Table~\ref{tab:conditioning} reports what doing so does to each model that can accept the other block: the change in test accuracy from $X$-only to cross-block conditioning, with every setting chosen on $X$-only validation.

\begin{table}[htbp]
\centering
\caption{What conditioning on the other outcome block does to each joint model, applied blindly as it must be in fusion: the change in test combined accuracy [pp] from $X$-only to cross-block conditioning, with the setting chosen on $X$-only validation. \emph{mean} is over all runs, \emph{worst} is the lowest cell mean, and \emph{cells} counts the grid cells in which the change is positive.}
\label{tab:conditioning}
\begin{tabular}{l rrr rrr}
\toprule
 & \multicolumn{3}{c}{Instacart} & \multicolumn{3}{c}{ACS-housing} \\
\cmidrule(lr){2-4} \cmidrule(lr){5-7}
model & mean & worst & cells & mean & worst & cells \\
\midrule
OBMP & $+0.188$ & $+0.089$ & 20/20 & $+0.355$ & $+0.031$ & 20/20 \\
MICE & $-2.028$ & $-5.135$ & 2/20 & $-4.628$ & $-7.135$ & 0/20 \\
factor analysis & $-0.585$ & $-1.056$ & 0/20 & $-0.441$ & $-0.751$ & 0/20 \\
MIWAE & $+0.027$ & $-0.095$ & 8/20 & $-0.853$ & $-2.654$ & 0/20 \\
MissForest$^{\dagger}$ & $-16.078$ & $-21.262$ & 0/20 & $-7.051$ & $-10.345$ & 0/20 \\
\bottomrule
\end{tabular}
\end{table}

OBMP gains in every cell of both grids: $+0.188$\,pp on Instacart and $+0.355$\,pp on ACS-housing on average, and $+0.089$ and $+0.031$\,pp in its worst cell. The joint imputers mostly lose. MICE loses $2.03$\,pp on Instacart and $4.63$\,pp on ACS-housing on average, and gains in only $2$ of the $40$ cells; factor analysis loses $0.59$ and $0.44$\,pp and gains in none; MissForest loses $16.1$ and $7.1$\,pp and gains in none. MIWAE is the one imputer that is not harmed on Instacart, gaining $+0.03$\,pp on average and in $8$ of $20$ cells, but on ACS-housing it loses $0.85$\,pp on average, $2.65$\,pp in its worst cell, and gains in none.

The joint imputers can accept the other block, but, trained without a single paired row, they mostly lose accuracy when they are given it, whereas OBMP gains in all $40$ cells. When the choice to condition cannot be validated, this is the property that matters: a model whose use of the other block does not hurt is one whose use of it can be decided in advance. 

\subsection{Decomposing the advantage}\label{sec:res-decomp}

Table~\ref{tab:decomposition} splits the total into two additive parts. Generative pre-training (OBMP $X$-only $-$ MLP) isolates what the generative stage adds over a randomly initialized network of the same capacity, trained on the same objective, for the same number of epochs, and selected on the same criterion. Cross-block conditioning (OBMP cross-block $-$ OBMP $X$-only) isolates what the joint model adds at inference time.

The two parts behave differently, and they behave differently in the same way on both datasets. Generative pre-training contributes $+0.046$\,pp on Instacart, resting almost entirely on the smallest sample size ($+0.193$\,pp at $n_{\mathrm{train}}{=}500$, $+0.042$\,pp at $n_{\mathrm{train}}{=}1000$, and within $0.004$\,pp of zero at every larger $n_{\mathrm{train}}$), and on ACS-housing it is absent altogether ($-0.016$\,pp over $100$ runs, $p = 0.053$): positive only at $n_{\mathrm{train}}{=}500$ ($+0.053$\,pp) and negative at every larger sample size ($-0.035$, $-0.028$, $-0.046$ and $-0.025$\,pp). As a way of initializing parameters, the generative stage buys nothing that enough epochs of discriminative training on a plain network would not also buy.

Cross-block conditioning is present in \emph{every cell of both grids}: $+0.188$\,pp on Instacart and $+0.355$\,pp on ACS-housing ($p < 0.001$), positive in all $40$ cells and, on ACS-housing, in $96$ of $100$ runs. It is what the total advantage over the discriminative arms is made of. OBMP is ahead of tuned $X$-logistic by $+0.293$ and $+0.438$\,pp and of the MLP by $+0.234$ and $+0.339$\,pp; the margin over the MLP is by construction the sum of the two terms, and on both datasets it consists mostly of the cross-block term.

The cross-block term deserves emphasis because of how it is measured. It compares one checkpoint against itself under two conditioning sets, so no training difference can contribute to it. And because $Y_A$ and $Y_B$ are never observed together in training, the association it exploits was never fitted from paired data; it is propagated through the shared hidden layers of the joint model. No model that maps $X$ to outcomes can produce this term, whatever its capacity.

\begin{table}[htbp]
\centering
\caption{The two contributions, as paired differences in test combined accuracy [pp], mean over five seeds. Positive favors OBMP. Cross-block conditioning is positive in every cell of both grids. The widest common block is $k=35$ on Instacart and $k=30$ on ACS-housing, whose $X$ has $30$ dimensions.}
\label{tab:decomposition}
\begin{tabular}{r cccc r cccc}
\toprule
 & pre- & cross- & vs. & vs. & & pre- & cross- & vs. & vs. \\
$k$ & training & block & MLP & $X$-log. & $k$ & training & block & MLP & $X$-log. \\
\midrule
\multicolumn{5}{l}{\emph{Instacart}} & \multicolumn{5}{l}{\emph{ACS-housing}} \\
\multicolumn{10}{l}{$n_{\mathrm{train}} = 500$} \\
5 & +0.380 & +0.186 & +0.566 & +0.311 & 5 & -0.018 & +0.061 & +0.043 & +0.045 \\
10 & +0.133 & +0.203 & +0.336 & +0.566 & 10 & +0.162 & +0.368 & +0.530 & +0.340 \\
20 & +0.159 & +0.141 & +0.299 & +0.444 & 20 & -0.015 & +0.300 & +0.285 & +0.487 \\
35 & +0.100 & +0.161 & +0.261 & +0.475 & 30 & +0.083 & +0.406 & +0.489 & +0.506 \\
\addlinespace
\multicolumn{10}{l}{$n_{\mathrm{train}} = 1{,}000$} \\
5 & +0.049 & +0.208 & +0.257 & +0.272 & 5 & -0.036 & +0.031 & -0.005 & +0.079 \\
10 & +0.034 & +0.273 & +0.307 & +0.430 & 10 & -0.082 & +0.458 & +0.377 & +0.413 \\
20 & +0.052 & +0.169 & +0.222 & +0.392 & 20 & +0.018 & +0.333 & +0.351 & +0.334 \\
35 & +0.033 & +0.134 & +0.167 & +0.366 & 30 & -0.039 & +0.383 & +0.345 & +0.405 \\
\addlinespace
\multicolumn{10}{l}{$n_{\mathrm{train}} = 2{,}000$} \\
5 & +0.014 & +0.129 & +0.143 & +0.163 & 5 & -0.026 & +0.089 & +0.063 & +0.198 \\
10 & -0.021 & +0.289 & +0.268 & +0.313 & 10 & +0.001 & +0.489 & +0.490 & +0.630 \\
20 & +0.025 & +0.167 & +0.192 & +0.242 & 20 & -0.047 & +0.362 & +0.315 & +0.346 \\
35 & -0.010 & +0.130 & +0.120 & +0.247 & 30 & -0.041 & +0.458 & +0.417 & +0.522 \\
\addlinespace
\multicolumn{10}{l}{$n_{\mathrm{train}} = 5{,}000$} \\
5 & +0.001 & +0.115 & +0.116 & +0.135 & 5 & +0.006 & +0.103 & +0.108 & +0.283 \\
10 & -0.004 & +0.266 & +0.262 & +0.264 & 10 & -0.059 & +0.568 & +0.509 & +0.724 \\
20 & +0.005 & +0.207 & +0.212 & +0.198 & 20 & -0.051 & +0.399 & +0.348 & +0.501 \\
35 & -0.010 & +0.179 & +0.169 & +0.231 & 30 & -0.077 & +0.514 & +0.437 & +0.519 \\
\addlinespace
\multicolumn{10}{l}{$n_{\mathrm{train}} = 10{,}000$} \\
5 & +0.011 & +0.089 & +0.099 & +0.111 & 5 & +0.009 & +0.171 & +0.180 & +0.414 \\
10 & -0.018 & +0.296 & +0.278 & +0.290 & 10 & -0.059 & +0.599 & +0.541 & +0.800 \\
20 & +0.003 & +0.205 & +0.207 & +0.204 & 20 & -0.034 & +0.491 & +0.458 & +0.633 \\
35 & -0.012 & +0.207 & +0.194 & +0.210 & 30 & -0.016 & +0.517 & +0.501 & +0.580 \\
\midrule
all & +0.046 & +0.188 & +0.234 & +0.293 & all & -0.016 & +0.355 & +0.339 & +0.438 \\
\bottomrule
\end{tabular}
\end{table}

\paragraph{The contrast is sensitive to how the checkpoint is chosen.} Tracking both checkpoints from each run (\S\ref{sec:exp-budget}) makes the size of that sensitivity explicit. On Instacart, under cross-block evaluation the cross-block-selected checkpoint scores $+0.026$\,pp higher than the $X$-only-selected one ($p = 0.002$), and reporting the contrast from it would inflate the cross-block term from $+0.188$ to $+0.230$\,pp. The inflation is not uniform: it reaches $+0.166$\,pp at $n_{\mathrm{train}}{=}10{,}000$, $k{=}5$ and is near zero at $k{=}35$, because a narrow common block leaves $X$-only validation least able to discriminate between candidate epochs. On ACS-housing the cross-block-selected checkpoint scores $+0.042$\,pp higher under cross-block evaluation ($p < 0.001$), with the largest difference at $n_{\mathrm{train}}{=}1000$, $k{=}10$ ($+0.099$\,pp). A decomposition that selects on the criterion it then reports would therefore attribute part of the selection procedure's behavior to the model. All numbers we report use the $X$-only-selected checkpoint.

\subsection{Is it really the association?}\label{sec:res-shuffle}

The account above says the cross-block term uses the dependence between the two outcome blocks. That is a causal claim about the data, and it can be tested directly by removing the dependence and watching the term go away. Permuting the rows of $Y_B$ does exactly that: every column marginal is preserved to the bit, $d_b$ and the base rates are unchanged, and only the correspondence with $Y_A$ is destroyed. Permuting a fraction $\alpha$ of the rows scales the damage. Table~\ref{tab:shuffle} reports the result on Instacart at $n_{\mathrm{train}} = 2000$, and the mean absolute between-block correlation confirms the manipulation does what it should, falling from $0.060$ to the $0.012$ finite-sample noise floor.

The cross-block term tracks it. From $+0.178$\,pp on untouched data it falls to $+0.063$, crosses zero near $\alpha = 0.5$, and settles at about $-0.03$\,pp once the association is gone (Spearman $\rho = -0.93$ over the $25$ seed means, $p < 10^{-10}$; the change from $\alpha=0$ to $\alpha=1$ is $-0.208$\,pp, larger than the term itself). The overshoot is what one would expect rather than a puzzle: conditioning on a block that carries no information about the target is not merely useless but costs a little, since the inference now has to accommodate values that explain nothing. All four values of $k$ show the same profile, positive at $\alpha=0$ and significantly negative at $\alpha=1$, so this is not one column of the grid driving the result.

Two features of the table make the reading unambiguous. First, generative pre-training --- the contribution that does not involve conditioning on $Y_B$ --- is flat across the whole sweep, within $\pm0.01$\,pp; the manipulation moves the term that should move and leaves alone the one that should not. Second, the collateral damage is small: permuting $Y_B$ also destroys $x \mapsto y_B$, but $X$ carries so little about $Y_B$ on this data that the $X$-only logistic arm loses only $0.048$\,pp on that block. Almost all of the $0.24$\,pp that OBMP gives up is the cross-block channel itself.

\begin{table}[htbp]
\centering
\caption{Breaking the association between the outcome blocks. Rows of $Y_B$ are permuted for a fraction $\alpha$ of the data, which leaves $d_b$, the base rates and every column marginal untouched and destroys only the dependence on $Y_A$; $\overline{|r|}$ is the resulting mean absolute correlation between the blocks on the test split. Instacart, $n_{\mathrm{train}}=2000$, all four $k$, five seeds [pp]. $^{*}$ marks $p<0.05$ against zero, clustered on seed.}
\label{tab:shuffle}
\begin{tabular}{c c cc c c}
\toprule
 & & \multicolumn{3}{c}{cross-block conditioning} & control \\
\cmidrule(lr){3-5} \cmidrule(lr){6-6}
$\alpha$ & $\overline{|r|}$ & combined & $Y_A$ & $Y_B$ & pre-training \\
\midrule
0.00 & 0.060 & $+0.178^{*}$ & $+0.288$ & $+0.052$ & $+0.002$ \\
0.25 & 0.048 & $+0.063^{*}$ & $+0.103$ & $+0.017$ & $+0.009^{*}$ \\
0.50 & 0.034 & $+0.002$ & $+0.003$ & $+0.002$ & $+0.005$ \\
0.75 & 0.019 & $-0.030^{*}$ & $-0.056$ & $-0.000$ & $+0.009$ \\
1.00 & 0.012 & $-0.029^{*}$ & $-0.055$ & $-0.000$ & $+0.005$ \\
\bottomrule
\end{tabular}
\end{table}

\subsection{Where the gain lands}\label{sec:res-blocks}

Table~\ref{tab:blocks} splits the cross-block term by outcome block, over the $k \in \{5, 10, 20\}$ the two grids share. On Instacart the gain concentrates on $Y_A$: $+0.326$\,pp there against $+0.046$ on $Y_B$, for $+0.196$ combined. On ACS-housing it is spread over both blocks: $+0.352$\,pp on $Y_A$ and $+0.282$ on $Y_B$, for $+0.322$ combined.

The difference follows how much room each block leaves. Instacart's $Y_B$ is near-saturated: tuned $X$-logistic does not beat a constant predictor on it ($91.2$\% against $91.3$\%), and conditioning on the other block buys little there. On ACS-housing both blocks have headroom (a constant predictor against tuned $X$-logistic, $79.5$\% and $82.0$\% on $Y_A$, $80.4$\% and $81.5$\% on $Y_B$), and both gain. The gain goes where there is room to gain, which is what one would expect if it comes from information the other block carries about the target, and not from how a particular dataset's blocks happen to be sized.

\begin{table}[htbp]
\centering
\caption{Cross-block conditioning by outcome block [pp], restricted to the $k \in \{5,10,20\}$ the two grids share. The combined figure is the dimension-weighted average \eqref{eq:combined} of the two block figures.}
\label{tab:blocks}
\begin{tabular}{l cc cc c}
\toprule
dataset & $d_a$ & $d_b$ & $Y_A$ & $Y_B$ & combined \\
\midrule
Instacart & 53 & 46 & $+0.326$ & $+0.046$ & $+0.196$ \\
ACS-housing & 22 & 17 & $+0.352$ & $+0.282$ & $+0.322$ \\
\bottomrule
\end{tabular}
\end{table}

\subsection{Trends in $n_{\mathrm{train}}$ and $k$}\label{sec:res-trends}

Table~\ref{tab:trends} regresses each contribution on $\log_2(n_{\mathrm{train}}/500)$ and $k-5$ with seed fixed effects. The sample-size axis is where the claim lies, and on it the cross-block term never decays. On Instacart its slope is $+0.0044$\,pp per doubling ($p = 0.32$, i.e.\ flat); on ACS-housing it is $+0.0380$ ($p < 0.05$, i.e.\ growing), and the term rises steadily, from $+0.284$\,pp at $n_{\mathrm{train}}{=}500$ through $+0.301$, $+0.350$ and $+0.396$ to $+0.445$\,pp at $n_{\mathrm{train}}{=}10{,}000$. Generative pre-training does not grow on either dataset: it falls at $-0.0390$\,pp per doubling on Instacart ($p = 0.03$) and has no significant slope on ACS-housing ($-0.0149$). The total advantage over $X$-logistic differs between the datasets because the cross-block term does. On Instacart, where that term is flat, the total falls at $-0.0585$ per doubling as the representation-based part of the margin shrinks; on ACS-housing, where the term grows, the total grows with it ($+0.0664$, $p < 0.05$). For a term whose value is the claim, not decaying across a twentyfold range of sample sizes is the result, and on one of the two datasets the term does more than hold its value.

The width axis gives no consistent picture. Instacart shows no significant $k$ slope ($-0.0010$, $p = 0.20$) and a non-monotone profile peaking at $k{=}10$. ACS-housing shows significantly positive slopes, $+0.0096$ for the cross-block term and $+0.0075$ for the total, but its profile is not monotone either: by $k$ the cross-block term is $+0.091$, $+0.497$, $+0.377$ and $+0.456$\,pp, with most of the rise between $k{=}5$ and $k{=}10$. We therefore make no claim about how the mechanism scales with the width of the common block. A study designed to answer it would need to vary $k$ while holding the composition of $X$ fixed.

\begin{table}[htbp]
\centering
\caption{Trend of each contribution in sample size and common-block width: OLS of the paired difference [pp] on $\log_2(n_{\mathrm{train}}/500)$ and $k-5$ with seed fixed effects. Coefficients come from the pooled fit; $^{*}$ marks $p<0.05$ from the same slope estimated within each seed and tested across the five.}
\label{tab:trends}
\begin{tabular}{ll rrr}
\toprule
dataset & contribution & $\beta_{\log_2 n_{\mathrm{train}}}$ & $\beta_k$ & $R^2$ \\
\midrule
Instacart & generative pre-training & $-0.0390^{*}$ & $-0.0016$ & 0.247 \\
 & cross-block conditioning & $+0.0044$ & $-0.0010$ & 0.070 \\
 & total vs.\ $X$-logistic & $-0.0585^{*}$ & $+0.0016$ & 0.524 \\
\addlinespace
ACS-housing & generative pre-training & $-0.0149$ & $-0.0004$ & 0.107 \\
 & cross-block conditioning & $+0.0380^{*}$ & $+0.0096^{*}$ & 0.378 \\
 & total vs.\ $X$-logistic & $+0.0664^{*}$ & $+0.0075^{*}$ & 0.350 \\
\bottomrule
\end{tabular}
\end{table}

\subsection{Does the mechanism need depth?}\label{sec:res-depth}

The model we have been calling a DBM has two hidden layers, and the mean-field sweep (\S\ref{sec:method-unroll}) carries a top-down term because of it. Table~\ref{tab:architecture} repeats both grids with that depth removed: $L = 1$ with $D_{h_1} = 96$, holding all of the DBM's hidden units in one layer, and $L = 1$ with $D_{h_1} = 64$, matching its first layer only. With one weight matrix the sweep loses its top-down term and the model is an RBM; nothing else changes.

\begin{table}[htbp]
\centering
\caption{Removing depth. The ablations replace the two hidden layers with one, keeping either the same total number of hidden units ($96$) or the first layer's ($64$). Accuracy and contrasts are means over the whole grid [\%, pp]. $^{*}$ marks a contrast differing from that dataset's DBM row at $p<0.05$, paired over cells.}
\label{tab:architecture}
\begin{tabular}{ll cc c}
\toprule
dataset & hidden layers & OBMP & cross-block & pre-training \\
\midrule
Instacart & $64 \to 32$ (DBM) & 83.815 & $+0.188$ & $+0.046$ \\
 & $96$ (RBM) & 83.843 & $+0.243^{*}$ & $+0.018^{*}$ \\
 & $64$ (RBM) & 83.797 & $+0.229^{*}$ & $-0.014^{*}$ \\
\addlinespace
ACS-housing & $64 \to 32$ (DBM) & 82.205 & $+0.355$ & $-0.016$ \\
 & $96$ (RBM) & 82.215 & $+0.325^{*}$ & $+0.024^{*}$ \\
 & $64$ (RBM) & 82.245 & $+0.362$ & $+0.018^{*}$ \\
\bottomrule
\end{tabular}
\end{table}

Depth is not what produces the effect. The cross-block term is present in all six configurations. On Instacart it is \emph{larger} without depth ($+0.243$ and $+0.229$\,pp against $+0.188$, both differing from the DBM at $p < 0.05$). On ACS-housing the $96$-unit RBM gives a slightly smaller term than the DBM ($+0.325$ against $+0.355$, a difference of $0.030$\,pp, $p = 0.019$) and the $64$-unit RBM one that does not differ from it significantly ($+0.362$). Every cell of every architecture on both datasets has a positive cross-block term. Removing the second hidden layer does not remove the term on either dataset, and it does not change the term in a consistent direction across the two: whatever the joint model is doing at inference time, it does not require a second hidden layer to do it.

On overall accuracy the architectures are indistinguishable for practical purposes: $83.815$, $83.843$ and $83.797$\% on Instacart for the DBM and the two RBMs, within $0.05$\,pp of one another, and $82.205$, $82.215$ and $82.245$\% on ACS-housing, within $0.04$\,pp. On neither dataset is the deeper model the most accurate. We keep the two-layer model as the main configuration because it is the one the decomposition was developed on, but the honest summary is that the second hidden layer is not carrying the result.

\paragraph{Nor more inference.} How long inference is allowed to run is a second thing one might expect the mechanism to depend on, since it is the one quantity here that can be spent freely at test time. Varying the cap over $T \in \{1, \ldots, 50\}$ on the primary checkpoints of both grids, it does not. The cross-block term is already significantly positive at $T = 1$ on both datasets (one pass is visible-to-hidden-to-visible, which is all an observed $Y_B$ needs to reach $Y_A$); on ACS-housing it is $+0.295$\,pp at $T = 1$ ($p < 0.001$, positive in all $20$ cells). It then settles onto a plateau, of $+0.18$ to $+0.23$\,pp on Instacart from $T = 10$ onward and of $+0.36$ to $+0.38$\,pp on ACS-housing from $T = 3$ onward ($+0.360$ at $T = 10$, $+0.355$ at $T = 15$, $+0.375$ at $T = 50$). On Instacart it converges no more slowly than the $X$-only term does, which is the opposite of what one would expect if the joint model's advantage had to be iterated into existence.

On neither dataset does the sweep converge within the cap for most runs. On ACS-housing only $12$ of the $100$ runs return at a cap of $25$ exactly what they return at $50$, yet the cross-block term stays on its plateau. On Instacart the tolerance is still not met at a cap of $50$ for most cells, so raising the cap keeps changing the predictions. What the extra passes buy there is absolute accuracy for \emph{both} conditioning sets, and slightly more of it for $X$-only, so the gap narrows rather than widens as inference proceeds. At $T \leq 2$ both Instacart arms score under a constant predictor, and $X$-only still does at $T = 3$ once $n_{\mathrm{train}} \geq 5000$, so the larger gaps in that region are between two failed predictors and we do not read them as the mechanism.

\section{Discussion}\label{sec:discussion}

\subsection{What the decomposition implies}\label{sec:disc-decomp}

Taken together, the results give a specific account of what a joint model contributes in data fusion. The generative criterion by itself is not competitive: ML-DBM is the weakest arm on both grids and on ACS-housing collapses onto the column-mean predictor in $99$ of $100$ runs. Discriminative fine-tuning through OBMP is what makes the model competitive. And what OBMP adds over a plain network of the same capacity is almost entirely the ability to condition on one outcome block when predicting the other, not the pre-trained initialization: the former is present in every cell of both datasets and does not decay in $n_{\mathrm{train}}$, the latter is confined to $n_{\mathrm{train}}{=}500$ on Instacart and absent on ACS-housing.

The practical consequence follows from which term survives. The contributions that come from representation shrink as the panels grow: the generative initialization, confined to $n_{\mathrm{train}}{=}500$ on either dataset, and on Instacart the total margin over a tuned discriminative model. That is the ordinary expectation for anything a flexible discriminative model can eventually learn on its own. The cross-block term does not shrink, and on ACS-housing it grows enough to carry the total margin over $X$-logistic up with it. Two things are being claimed there and they have different standing. That the term cannot be reproduced by a discriminative model is structural: such a model has no argument to put $y_B$ into, whatever its capacity or sample size. That the term is \emph{positive} is empirical, and rests on the conditional independence assumption (\S\ref{sec:method-instantiation}) holding well enough on the data at hand. \S\ref{sec:disc-size} sets these magnitudes against what the task allows, and \S\ref{sec:res-depth} shows the term does not depend on the model being deep.

The comparison with tuned imputers adds a practical argument for the joint model that the decomposition alone does not make. Several imputers can accept the other outcome block as evidence, and on a given dataset one of them may use it well. But a practitioner with fusion data cannot tell which case they are in, since no row observes both blocks, and so has to decide whether to condition without being able to validate the decision (\S\ref{sec:res-conditioning}). What matters then is not how much a model can gain from the other block where it happens to help, but whether conditioning can hurt. For the joint imputers here it usually does, by up to several points; for OBMP it did not reduce accuracy in any cell of either grid. That is the property a method needs when the choice has to be made blind.

\subsection{The size of the effect}\label{sec:disc-size}

The margins reported here are small in absolute terms, and it is worth being explicit about what they are small relative to. Both tasks admit little movement. On Instacart, predicting zero everywhere already scores $81.3$\% combined accuracy and the strongest tuned baseline on each run reaches $83.7$\%, so the distance that baseline covers from the constant predictor is $2.4$ points; OBMP's $+0.13$\,pp over it is $5$\% of that distance. On ACS-housing a constant predictor scores $79.9$\% and the strongest tuned baseline $81.8$\%, a distance of $1.9$ points, of which OBMP's $+0.41$\,pp is $22$\%. On cross-entropy, where the ceiling does not bind in the same way, the reduction against $X$-logistic is $2.9$\% on Instacart and $3.3$\% on ACS-housing.

A logistic regression fitted on complete rows also gives a rough ceiling for the cross-block term. With paired rows available, which fusion data never have, that regression gains $0.13$\,pp on Instacart and $1.32$\,pp on ACS-housing from adding the other outcome block, at $n_{\mathrm{train}}{=}2000$ and the widest $k$. OBMP, which never sees a paired row, gains $0.13$ and $0.46$\,pp from the other block in those cells. Instacart's term is at this ceiling; ACS-housing's is about a third of it, so on that dataset there is room for a method that uses the other block more fully.

The two outcome blocks also differ in how far they can be predicted at all. On Instacart's $Y_A$ the constant predictor scores $72.7$\% and tuned $X$-logistic $76.7$\%; on $Y_B$ they score $91.3$\% and $91.2$\%, that is, the baseline does not beat a constant on $Y_B$. On ACS-housing both blocks have headroom: the same two predictors score $79.5$\% and $82.0$\% on $Y_A$, and $80.4$\% and $81.5$\% on $Y_B$. Averaging the blocks by dimension, as \eqref{eq:combined} does, therefore dilutes the contrasts on Instacart more than on ACS-housing (\S\ref{sec:res-blocks}). We report the diluted figure throughout because it is the honest summary of the task as posed, but a reader setting these numbers against gains reported on tasks with more headroom should keep the difference in view.

None of this makes the effect large, and the argument of this paper does not rest on its size. It rests on which part of it survives. A margin that shrinks with $n_{\mathrm{train}}$ is one a discriminative model will close given enough data, and the representation-based part of what we measure behaves that way. The cross-block term does not, on either dataset, and the reason is structural: it uses evidence that a model mapping $X$ to outcomes has no way to accept. A small effect that does not decay is a different kind of finding from a small effect that does, and it is the first kind we claim here.

\subsection{Limitations}\label{sec:limitations}

\paragraph{Three baselines do not emit probabilities.} Hot deck, propensity matching and MissForest return hard values; MissForest in particular scores below a constant zero predictor because a regression forest's continuous output is thresholded. Their accuracy gaps overstate the methodological distance and their cross-entropies are not comparable.

\paragraph{The $k$ axis is confounded.} Training subsets are drawn with a generator seeded jointly on the seed and $k$, so cells that differ in $k$ use different rows, and the $k$ trends in Table~\ref{tab:trends} are estimated across that extra variation. More importantly, $X$ is restricted to width $k$ by a popularity ranking, so increasing $k$ adds progressively less common covariates (aisles further down the purchase ranking on Instacart, rarer demographic categories on ACS-housing), and the $k$ slope conflates ``more covariates'' with ``sparser covariates''. It also changes which kinds of covariate $X$ holds: on ACS-housing the popularity ranking brings all age bands into $X$ at $k{=}20$. This is the most likely reason the $k$ profiles differ between the datasets and are not monotone (\S\ref{sec:res-trends}), and it is why we make no claim on that axis.

\paragraph{Panels split rather than collected.} Both grids are built by partitioning one collected table into blocks, which is standard practice for evaluating fusion but is not the situation the method is for: in a real fusion the two panels come from different instruments, with different measurement error, coverage and definitions, and nothing guarantees the shared covariates mean the same thing in both. Our outcomes are also all binary. The mechanism should be tested on genuinely separate panels and on continuous outcomes before the magnitudes here are taken as representative.

\paragraph{Two backends.} The model arms were run on Apple MPS for Instacart and on CPU for ACS-housing; the baselines were run on CPU for both. Because the persistent chains of the generative stage draw from the device's random number generator, a change of backend moves the arms that start from it slightly; the calibration in Appendix~\ref{app:repro} puts that movement far below the effects reported here. Each comparison between methods is made within one dataset, where every model arm was run on the same backend.

\paragraph{Finite tuning grids.} The baselines are tuned over modest grids (Appendix~\ref{app:repro}), selected on $X$-only validation. A practitioner with different grids could do better with them than we did. The crediting rule of Table~\ref{tab:main-accuracy}, which lets each joint imputer use whichever conditioning set scores better on the test split, already leans in the baselines' favour.

\section{Conclusion}\label{sec:conclusion}

Data fusion withholds the one thing a discriminative criterion needs: a row where the quantity to be predicted is recorded next to the evidence for it. Restricting multi-prediction training to targets the data actually contains removes that obstacle, and the resulting criterion is defined for any missingness pattern while reducing to the original one when rows are complete. That is a modest technical step, but it is what makes the substantive question askable, because it puts a discriminatively fine-tuned joint model and a discriminative model of the same capacity on the same footing.

Asked that way, the answer is narrower than the totals suggest. Against baselines tuned on validation and given the same conditioning, the fine-tuned DBM is the best of all methods in $37$ of $40$ cells, but the generative stage it starts from contributes almost nothing: it is confined to the smallest sample size on one dataset and absent on the other, and removing the second hidden layer does not remove the effect either. What the joint model supplies is the ability to condition on one outcome block while predicting the other, worth $+0.19$ and $+0.36$\,pp, positive in all $40$ cells, and alone among the contributions in not fading as the panels grow, as the architecture is flattened, or as inference is allowed to run longer. The imputers that can also accept the other block mostly lose accuracy when they do, and since fusion data cannot validate that choice, a model whose use of the other block does not hurt is what a practitioner needs. The effect is small, and it is the part a discriminative model cannot reach at any capacity or sample size, because the evidence it uses is not of the form such a model accepts. How the mechanism scales with the width of the common block is not consistent across our datasets, and we make no claim on that axis. Testing the mechanism on genuinely separate panels, collected by different instruments rather than split from one table, is what we would do next.

\section*{Acknowledgment}
This study is supported by JSPS KAKENHI (Grant No. JP24K\allowbreak{}16472).

\appendix

\section{Implementation details}\label{app:repro}

Every arm is trained per cell with the seed fixed to the cell's seed, so on a given backend a cell is reproducible from its $(n_{\mathrm{train}}, k, \mathrm{seed})$ alone. The derived ACS-housing data are released as described in \S\ref{sec:exp-data}.

\paragraph{Stage one (ML-DBM).} SGD, learning rate $0.005$, batch size $64$, $\max(15, \lceil 25000/n_{\mathrm{train}}\rceil)$ epochs. The positive phase uses $15$ mean-field passes; the negative phase is persistent contrastive divergence with a chain of $64$ states carried across minibatches and $5$ Gibbs steps per update. With $p_{\mathrm{complete}}=0$ there is no complete subset to pre-train layerwise on, so we skip greedy pre-training and initialize the visible bias to $\mathrm{logit}(\bar v_j)$ over the entries with $m_{ij}=1$, clamped to $[10^{-3}, 1-10^{-3}]$.

\paragraph{Stage two (OBMP).} AdamW, learning rate $10^{-3}$, no weight decay, batch size $64$, gradient-norm clipping at $5.0$, up to $200$ epochs with validation every $5$ and patience $60$. The differentiable unroll runs a fixed $T=10$ passes with no damping.

\paragraph{MLP.} The same optimizer, learning rate, batch size, epoch budget, evaluation cadence and patience as stage two, with weight decay selected on validation from $\{0, 10^{-4}, 10^{-3}, 10^{-2}\}$. Architecture $k \to 64 \to 32 \to (d_a + d_b)$ with sigmoid activations, matching the DBM's hidden widths. The loss is the observed-entry cross-entropy normalized within each row before averaging over rows, so that it matches \eqref{eq:obmp} rather than weighting rows by how many targets they contribute.

\paragraph{Tuned $X$-logistic.} One \texttt{liblinear} logistic regression per outcome dimension, $\texttt{max\_iter}=500$, inverse regularization strength selected on validation from $\{10^{-3}, 10^{-2}, 10^{-1}, 1, 10\}$. A dimension with fewer than ten observed training rows, or with only one class among them, falls back to the observed marginal.

\paragraph{Baselines.} All baselines are fitted on the same training rows as the model arms. Those with free parameters are tuned per cell on $X$-only validation combined accuracy, with ties broken by cross-entropy, over the following grids: $k$-NN with $k \in \{5, 15, 50, 150\}$; MICE as \texttt{IterativeImputer} with a ridge solver (the library default, \texttt{BayesianRidge}, raised a LAPACK convergence failure on every cell we tried), ridge strength $\alpha \in \{0.1, 1, 10, 100\}$ and $\{3, 10\}$ cycles; factor analysis with $\{2, 4, 8, 16, 32\}$ components; MissForest as \texttt{IterativeImputer} with a random-forest regressor and $3$ cycles, with (trees, depth) $\in \{(10, 8), (50, 8), (10, 10)\}$; MIWAE with $128$ hidden units, $10$ importance samples in training and $50$ at evaluation, batch $256$ and learning rate $10^{-3}$, with latent dimension in $\{8, 16, 32\}$ and the epoch chosen every $20$ epochs up to $200$. Tuned $X$-logistic uses the grid given above. Hot deck, propensity-score matching, the column mean and the mode-zero predictor have nothing to tune. Each grid contains the fixed setting the baseline was previously run at, and on the machine that ran the ACS-housing cells every such setting reproduces the untuned result exactly; tuning alone raised the baselines by $0.2$ to $2.8$\,pp over those fixed settings, measured on a subset of the ACS-housing cells. The four joint imputers (MICE, factor analysis, MissForest and MIWAE) are evaluated both with the other outcome block hidden and with it supplied as evidence. In both cases the setting is the one chosen on $X$-only validation, since fusion data offer no row on which a setting for cross-block conditioning could be validated.

\paragraph{Data preparation.} Instacart is reduced to a user $\times$ aisle purchase-indicator matrix over $134$ aisles; users are sampled without replacement from the full pool by a seeded permutation and sliced into train, validation and test. ACS-housing is built from the California person and housing records of the 2024 ACS 1-year PUMS: each occupied housing unit contributes its householder, aged 18 or over, joined to the housing record, and group quarters are excluded. Categorical fields are one-hot encoded, wages are binned to tertiles, and personal and household income to quartiles; survey weights are not used. Block membership is fixed by subject matter as described (\S\ref{sec:exp-data}); within a cell, the training subset of size $n_{\mathrm{train}}$ is drawn by a permutation seeded on $(\mathrm{seed}, k)$, which makes subsets nested in $n_{\mathrm{train}}$ within a seed but not aligned across $k$ (\S\ref{sec:limitations}).

\paragraph{Evaluation.} Test-time mean-field runs to a cap of $T=15$ passes with early exit once no coordinate moves by more than $10^{-4}$. Predictions are thresholded at $0.5$ for accuracy and used as probabilities for cross-entropy.

\paragraph{Backends.} The model arms were run on Apple MPS for Instacart and on CPU (x86, one thread per process) for ACS-housing; all baselines were run on CPU. The persistent chains of stage one draw from the device's random number generator, so a change of backend moves the arms that depend on them. Rerunning the Instacart cells at $n_{\mathrm{train}}{=}2000$ and the widest $k$ on CPU moved OBMP by less than $0.01$\,pp on average and the cross-block term by at most $0.003$\,pp, while $X$-logistic and the MLP are bit-identical across backends.

\paragraph{Reproducibility of the baselines.} $k$-NN does not reproduce exactly across machines. On binary data many Hamming distances tie, and how the partial sort breaks ties depends on the build of the numerical library; the Instacart $k$-NN results differed by up to $2.1$\,pp between two machines. MIWAE differs by up to $0.6$\,pp. MICE and MissForest reproduce exactly.

\bibliographystyle{unsrt}
\bibliography{../mldbm}

\begin{thebibliography}{10}

\bibitem{bm}
Geoffrey Hinton.
\newblock Nobel lecture: Boltzmann machines.
\newblock {\em Reviews of Modern Physics}, 97(3):030502, 2025.

\bibitem{bm_training}
David~H Ackley, Geoffrey~E Hinton, and Terrence~J Sejnowski.
\newblock A learning algorithm for boltzmann machines.
\newblock {\em Cognitive science}, 9(1):147--169, 1985.

\bibitem{ebt}
Alexi Gladstone, Ganesh Nanduru, Md~Mofijul Islam, Peixuan Han, Hyeonjeong Ha,
  Aman Chadha, Yilun Du, Heng Ji, Jundong Li, and Tariq Iqbal.
\newblock Energy-based transformers are scalable learners and thinkers.
\newblock In {\em The Fourteenth International Conference on Learning
  Representations}, 2026.

\bibitem{i-jepa}
Mahmoud Assran, Quentin Duval, Ishan Misra, Piotr Bojanowski, Pascal Vincent,
  Michael Rabbat, Yann LeCun, and Nicolas Ballas.
\newblock Self-supervised learning from images with a joint-embedding
  predictive architecture.
\newblock In {\em IEEE/CVF Conference on Computer Vision and Pattern
  Recognition}, 2023.

\bibitem{v-jepa}
Adrien Bardes, Quentin Garrido, Jean Ponce, Xinlei Chen, Michael Rabbat, Yann
  LeCun, Mahmoud Assran, and Nicolas Ballas.
\newblock Revisiting feature prediction for learning visual representations
  from video.
\newblock {\em arXiv preprint arXiv:2404.08471}, 2024.

\bibitem{lecun2022}
Yann LeCun.
\newblock A path towards autonomous machine intelligence.
\newblock OpenReview \url{https://openreview.net/forum?id=BZ5a1r-kVsf}, 2022.

\bibitem{ebm_hitchhiker}
Davide Carbone.
\newblock Hitchhiker's guide on the relation of energy-based models with other
  generative models, sampling and statistical physics: a comprehensive review.
\newblock {\em Transactions on Machine Learning Research}, 2025.

\bibitem{ebm_tutorial}
Yann LeCun, Sumit Chopra, Raia Hadsell, M~Ranzato, Fujie Huang, et~al.
\newblock A tutorial on energy-based learning.
\newblock {\em Predicting structured data}, 2006.

\bibitem{ebm_train}
Yang Song and Diederik~P Kingma.
\newblock How to train your energy-based models.
\newblock {\em arXiv preprint arXiv:2101.03288}, 2021.

\bibitem{ebm_neurips2025}
Louis B{\'e}thune, David Vigouroux, Yilun Du, Rufin VanRullen, Thomas Serre,
  and Victor Boutin.
\newblock Follow the energy, find the path: Riemannian metrics from
  energy-based models.
\newblock {\em Advances in Neural Information Processing Systems},
  38:97824--97870, 2026.

\bibitem{dbm}
Ruslan Salakhutdinov and Geoffrey Hinton.
\newblock Deep boltzmann machines.
\newblock In {\em Artificial intelligence and statistics}, pages 448--455.
  PMLR, 2009.

\bibitem{kamakura_wedel_1997}
Wagner~A Kamakura and Michel Wedel.
\newblock Statistical data fusion for cross-tabulation.
\newblock {\em Journal of Marketing Research}, 34(4):485--498, 1997.

\bibitem{niimi_hoshino_2017}
Junichiro Niimi and Takahiro Hoshino.
\newblock A method for data fusion using deep boltzmann machine: An application
  of dbm in data fusion to predict customers behavior at competitors.
\newblock {\em Proceedings of the Annual Conference of JSAI}, page 1I12, 2017.

\bibitem{mp-dbm}
Ian Goodfellow, Mehdi Mirza, Aaron Courville, and Yoshua Bengio.
\newblock Multi-prediction deep boltzmann machines.
\newblock {\em Advances in Neural Information Processing Systems}, 26, 2013.

\bibitem{smolensky_rbm}
Paul Smolensky.
\newblock Information processing in dynamical systems: Foundations of harmony
  theory.
\newblock Technical report, Department of Computer Science, University of
  Colorado at Boulder, 1986.

\bibitem{hinton_rbm}
Geoffrey~E Hinton.
\newblock A practical guide to training restricted boltzmann machines.
\newblock In {\em Neural Networks: Tricks of the Trade: Second Edition}, pages
  599--619. Springer, 2012.

\bibitem{salakhutdinov_murray_2008}
Ruslan Salakhutdinov and Iain Murray.
\newblock On the quantitative analysis of deep belief networks.
\newblock In {\em Proceedings of the 25th international conference on Machine
  learning}, pages 872--879, 2008.

\bibitem{propensityscore}
Paul~R Rosenbaum and Donald~B Rubin.
\newblock The central role of the propensity score in observational studies for
  causal effects.
\newblock {\em Biometrika}, pages 41--55, 1983.

\bibitem{clustering_fusion}
Bailey~K Fosdick, Maria DeYoreo, Jerome~P Reiter, et~al.
\newblock Categorical data fusion using auxiliary information.
\newblock {\em The Annals of Applied Statistics}, 10(4):1907--1929, 2017.

\bibitem{mice}
Stef Van~Buuren and Karin Groothuis-Oudshoorn.
\newblock mice: Multivariate imputation by chained equations in r.
\newblock {\em Journal of statistical software}, 45:1--67, 2011.

\bibitem{missforest}
Daniel~J Stekhoven and Peter B{\"u}hlmann.
\newblock Missforest---non-parametric missing value imputation for mixed-type
  data.
\newblock {\em Bioinformatics}, 28(1):112--118, 2012.

\bibitem{vae}
Diederik~P. Kingma and Max Welling.
\newblock Auto-encoding variational bayes.
\newblock In {\em Proceedings of the 2nd International Conference on Learning
  Representations (ICLR 2014)}, 2014.

\bibitem{miwae}
Pierre-Alexandre Mattei and Jes Frellsen.
\newblock Miwae: Deep generative modelling and imputation of incomplete data
  sets.
\newblock In {\em International conference on machine learning}, pages
  4413--4423. PMLR, 2019.

\bibitem{multimodal_dbm}
Nitish Srivastava and Russ~R Salakhutdinov.
\newblock Multimodal learning with deep boltzmann machines.
\newblock {\em Advances in neural information processing systems}, 25, 2012.

\bibitem{rubin1976}
Donald~B Rubin.
\newblock Inference and missing data.
\newblock {\em Biometrika}, 63(3):581--592, 1976.

\bibitem{raessler2004}
Susanne R{\"a}ssler.
\newblock Data fusion: Identification problems, validity, and multiple
  imputation.
\newblock {\em Austrian Journal of Statistics}, 33(1--2):153--171, 2004.

\end{thebibliography}

\end{document}